%% file: main.tex
\documentclass[10pt,twocolumn,letterpaper]{article}

\usepackage[pagenumbers]{iccv} 
\usepackage[para]{footmisc}

\usepackage{multirow}
\usepackage[table]{xcolor}
\usepackage{algpseudocode}
\usepackage{overpic}
\usepackage[T1]{fontenc}

\definecolor{iccvblue}{rgb}{0.21,0.49,0.74}
\usepackage[pagebackref,breaklinks,colorlinks,allcolors=iccvblue]{hyperref}

\usepackage{algpseudocode}
\usepackage[ruled]{algorithm2e}
\usepackage{float}
\usepackage{listings}
\usepackage{xcolor}
\usepackage{graphbox}
\usepackage{placeins}

\lstdefinestyle{prompt}{
    basicstyle=\ttfamily\small,  
    backgroundcolor=\color{blue!5}, 
    frame=single,                
    framerule=1pt,             
    rulecolor=\color{blue!10},      
    breaklines=true,             
    keepspaces=true,             
    columns=flexible,            
    breakautoindent=false,       
    breakindent=0.1pt,
    breakatwhitespace=true,
}

\makeatletter
\long\def\@makefntext#1{\leavevmode
  \@makefnmark\nobreak
  #1%
}
\makeatother

\def\paperID{00006} 
\def\confName{ICCV}
\def\confYear{2025}

\title{Sketch-to-Layout: Sketch-Guided Multimodal Layout Generation}
\author{
    Riccardo Brioschi\thanks{These authors contributed equally and are listed in a \href{https://www.aeaweb.org/journals/policies/random-author-order/search?RandomAuthorsSearch\%5Bsearch\%5D=Sketch-to-Layout\%3A+Sketch-Guided+Multimodal+Layout+Generation}{certified random order.}}\hspace{2mm}\thanks{EPFL, work done during time as Student Researchers at Google DeepMind.} \\
    \and
    Aleksandr Alekseev\textsuperscript{*}\footnotemark[2] \\
    \and
    Emanuele Nevali\textsuperscript{*}\footnotemark[2] \\
    \and
    Berkay Döner\textsuperscript{*}\footnotemark[2] \\
    \and
    Omar El Malki\textsuperscript{*}\footnotemark[2] \\
    \and
    Blagoj Mitrevski\thanks{Google DeepMind. Correspondence to:
Blagoj Mitrevski \href{mailto:bmitrevski@google.com}{\textless bmitrevski@google.com\textgreater}} \\
    \and
    Leandro Kieliger\footnotemark[3] \\
    \and
    Mark Collier\footnotemark[3] \\
    \and
    Andrii Maksai\footnotemark[3] \\
    \and
    Jesse Berent\footnotemark[3] \\
    \and
    Claudiu Cristian Musat\footnotemark[3] \\
    \and
    Efi Kokiopoulou\footnotemark[3]
}

\begin{document}
\maketitle
\input{sec/0_abstract}
\input{sec/1_intro}
\input{sec/2_related_work}
\input{sec/3_perliminary_experiments}
\input{sec/4_methodology}
\input{sec/5_experiments}
\input{sec/6_conclusion}
{
    \small
    \bibliographystyle{ieeenat_fullname}
    \bibliography{main}
}

\input{sec/7_suppl}

\end{document}

%% file: sec/0_abstract.tex
\begin{abstract}
Graphic layout generation is a growing research area focusing on generating aesthetically pleasing layouts ranging from poster designs to documents. While recent research has explored ways to incorporate user constraints to guide the layout generation, these constraints often require complex specifications which reduce usability. We introduce an innovative approach exploiting user-provided sketches as intuitive constraints and we demonstrate empirically the effectiveness of this new guidance method, establishing the sketch-to-layout problem as a promising research direction, which is currently under-explored. To tackle the sketch-to-layout problem, we propose a multimodal transformer-based solution using the sketch and the content assets as inputs to produce high quality layouts. Since collecting sketch training data from human annotators to train our model is very costly, we introduce a novel and efficient method to synthetically generate training sketches at scale. We train and evaluate our model on three publicly available datasets: PubLayNet \cite{zhong2019publaynetlargestdatasetdocument}, DocLayNet \cite{PfitzmannDoclaynet2022} and SlidesVQA \cite{tanaka2023slidevqadatasetdocumentvisual}, demonstrating that it outperforms state-of-the-art constraint-based methods, while offering a more intuitive design experience.
In order to facilitate future sketch-to-layout research, we release O(200k) synthetically-generated sketches for the public datasets above.\footnote{The datasets are available at \url{https://github.com/google-deepmind/sketch_to_layout}}
\end{abstract}

%% file: sec/1_intro.tex
\begin{figure}[tbh]
\centering
     \includegraphics[width=0.4\textwidth]{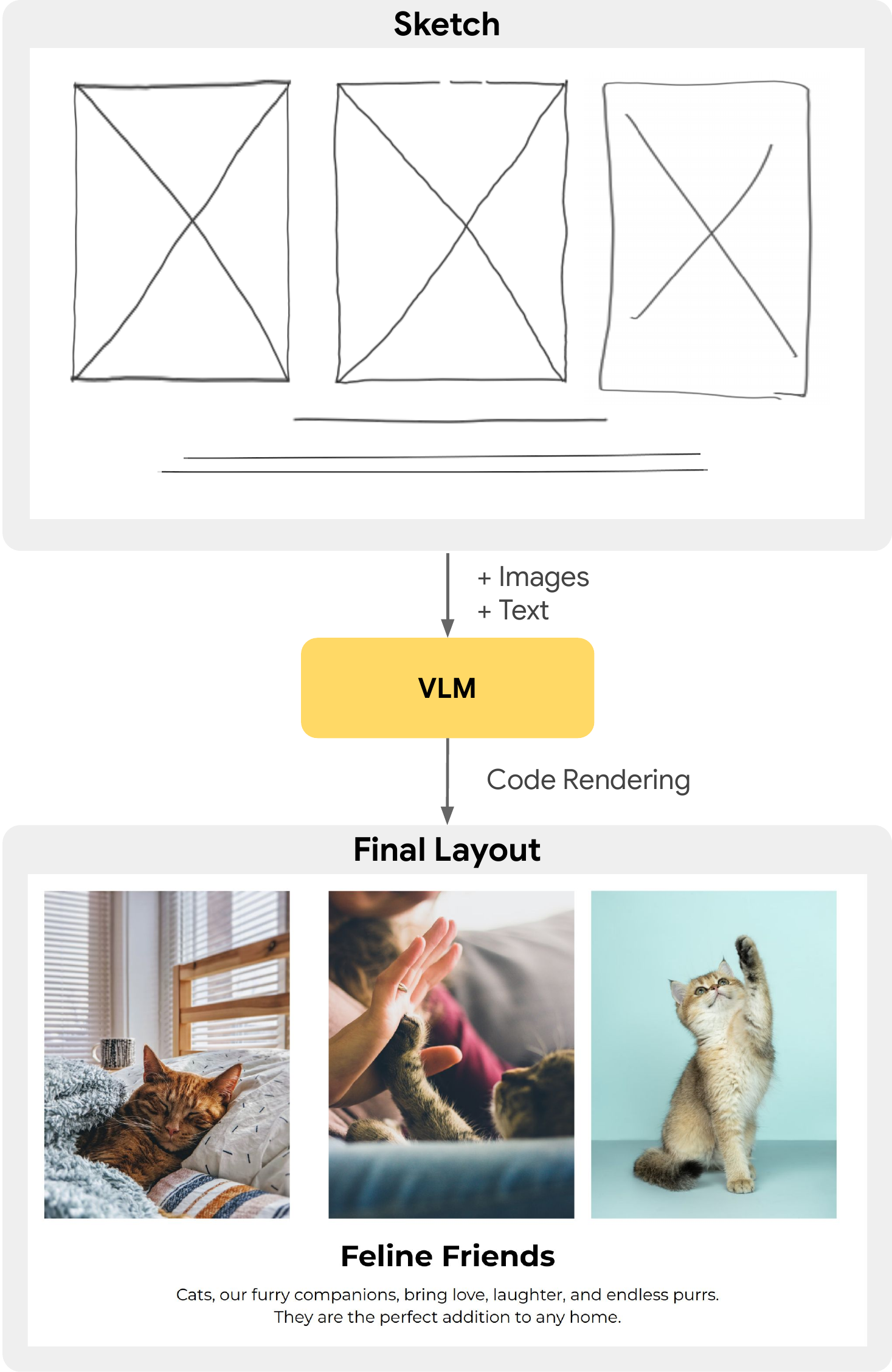}
     \caption{Our sketch-to-layout approach leverages sketches to guide the generation of multimodal layouts in a natural and intuitive way.}
     \label{fig:cover_figure_vertical}
\end{figure}

\section{Introduction}
Designing aesthetically pleasing and usable layouts for graphic design is a fundamental challenge. Layouts should represent a visually pleasing arrangement of text and image elements with appropriate sizes and positions, while at the same time capturing the right information hierarchy. Assets should have consistent semantic relationships such as an engaging reading order. Manual design can be time-consuming and automated layout generation aims to reduce this burden.

Recent research has explored various approaches to layout generation, including image generation methods such as GANs \cite{li2020layoutgan,Kikuchi_2021} and LLM-based methods \cite{tang2023layoutnuwa, lin2024layoutprompter, yang2024posterllava}. Some of these approaches try to incorporate user-defined constraints to guide the generation, but require complex details such as specifying precise element dimensions  \cite{lin2024layoutprompter, jiang2023layoutformer++}, complex positional relationships \cite{lin2024layoutprompter, jiang2023layoutformer++,yang2024posterllava,Kikuchi_2021}, grid-based guidelines \cite{cheng2023playparametricallyconditionedlayout} and detailed textual descriptions \cite{lin2023parse}. It is cumbersome for users to come up with such complex constraints. On the contrary, sketching, a common design practice where users quickly outline the structure of a design, offers a more intuitive alternative. Sketching is a widely used technique for creative tasks that captures the high-level essence of a layout without requiring overwhelming detail. Studies on designer behavior \cite{myers2008designers,newman2000sitemaps} show that sketching is an integral starting point for almost all designers in various domains \cite{buxton2010sketching}.

This paper proposes using sketches as intuitive constraints to guide the generation of multimodal (image and text) layouts. We start by demonstrating empirically the effectiveness of sketches as a new guidance method when compared against other forms of user-defined constraints. This empirical result suggests that the sketch-to-layout problem is a promising direction for constrained layout generation, which has been under-explored. To tackle the sketch-to-layout problem, we propose a solution leveraging Vision-Language Models (VLMs); see also Fig \ref{fig:cover_figure_vertical}.
Our method takes as input a user constraint in a form of a sketch along with image and text assets, to produce visually pleasing high quality layouts, capturing the structure suggested by the user while maintaining aesthetic appeal. 

Although VLMs have shown impressive performances on a wide range of tasks, recent research \cite{li2024sketch2code} demonstrated that generating the correct layout from sketch inputs in a single step is challenging even for state-of-the-art VLMs, which amplifies the need of collecting sketch training data for improving performance. However, collecting sketch data from human annotators to train a VLM model is very costly and time-consuming. In order to address this challenge, we propose a novel and efficient technique to synthetically generate sketches at scale, unblocking the fine-tuning of VLMs to tackle the sketch-to-layout problem.
In order to accelerate research progress on sketch-to-layout, we release a dataset of O(200k) synthetic sketches generated by our proposed method.

Our VLM-based approach is general and applicable to any VLM. In our empirical study we use PaLIGemma 3B \cite{beyer2024paligemmaversatile3bvlm} as an example open-source VLM. We train and evaluate our model on three publicly available datasets: PubLayNet \cite{zhong2019publaynetlargestdatasetdocument}, DocLayNet \cite{PfitzmannDoclaynet2022} and SlidesVQA \cite{tanaka2023slidevqadatasetdocumentvisual}, demonstrating that it outperforms state-of-the-art constraint-based methods, while offering a more intuitive design experience. 
By evaluating our model on both synthetic and human-produced sketches, we arrive at comparable performance, which validates the use of synthetic sketches as a reliable proxy for actual human-produced sketches when used as training data for VLMs in the sketch-to-layout task. In summary, our contributions are as follows:

\begin{itemize}
    \item We demonstrate the value of sketches as a novel guidance method for layout generation, establishing the sketch-to-layout as an effective research direction for guided layout generation.
    \item We introduce a novel methodology to create large-scale synthetic datasets with sketches of documents and layouts, unblocking efficient VLM training and evaluation.
    \item We release our large collection of O(200k) synthetically generated sketches for three publicly available datasets, in order to facilitate future research in this previously data-scarce domain.
    \item We provide experimental results showing that our method leveraging PaLIGemma outperforms state-of-the-art constraint-based layout generation methods by more than 40\% in terms of Maximum IoU on three publicly available datasets. Our empirical results also highlight the importance of the content-awareness aspect of our method.
    \item We introduce Content Ordering Score (COS), a new metric inspired by the order loss \cite{li2020attributeconditionedlayoutganautomatic}, designed to assess the content-awareness of a generated layout.

\end{itemize}

%% file: sec/2_related_work.tex
\section{Related Work}

\textbf{Unconstrained Generation.}
Early research in layout generation primarily focused on unconditional generation. CanvasVAE \cite{yamaguchi2021canvasvae} models documents as a combination of canvases and elements, adopting a VAE to capture the distribution of their attributes. \citet{gupta2021layouttransformer} propose an auto-regressive transformer to frame layout generation as a sequence-to-sequence task, and show the effectiveness of their approach on different domains.

\textbf{Constrained Generation.}
We focus on constrained generation of layouts, which has been investigated before with different types of constraints. LayoutVAE \cite{jyothi2019layoutvae} proposes a two-stage VAE model that takes the set of labels as input constraint. Similarly, LayoutGAN \cite{li2020layoutgan} synthesizes layouts given the set of labels with each label having a separate probability distribution in the generator. Further work from \citet{li2020attributeconditionedlayoutganautomatic} uses the area, aspect ratio and reading-order of the input elements as the input constraints. \cite{lee2020neural} takes relational constraints between elements (such as specifying a text block to be on the left of an image) and models the relationships using a graph-based model. \cite{Kikuchi_2021} uses a transformer-based GAN that can additionally take beautification constraints, such as alignment and non-overlap. Later work using diffusion models \cite{Inoue_2023_CVPR, cheng2023playparametricallyconditionedlayout}, transformers \cite{jiang2023layoutformer++}, and LLMs \cite{lin2024layoutprompter} focus on respecting different constraints including layout grids,  category types and sizes of elements and the relationships between them. Recent work incorporates textual descriptions of layouts to guide the generation \cite{lin2023parse, lin2024layoutprompter}. While textual descriptions are a step towards more intuitive constraints compared to the previously adopted ones, we argue that sketches offer an even more direct and natural way for users to express their layout preferences.

\textbf{Content-Aware Methods.} Some prior work has incorporated content in the form of user-provided assets to guide the generation. \citet{zheng-sig19} take images, keywords and the layout category as input, which are fed into separate encoders guiding a layout generation GAN. CGL-GAN \cite{zhou2022composition} takes the background canvas as input and leverages its saliency map to guide the generation process. Later work in the area also focused on placing elements on a provided background using different architectures, such as CNN-LSTM-based-GAN \cite{hsu2023posterlayout} and pretrained VLMs \cite{seol2024posterllama, yang2024posterllava}. 
\citet{yu2022layoutdetr} uses an object detection transformer model (DETR\cite{carion2020end}) to guide the generation, relying on ViT\cite{dosovitskiy2020image} and BERT \cite{devlin2019bertpretrainingdeepbidirectional} to encode input images and texts.
\cite{inoue2023towards} considers documents as a set of multimodal elements and uses CLIP to embed textual and visual features. Similarly, \cite{shabani2024visual} utilizes CLIP embeddings for input elements and vector attributes as conditions to diffusion model. Similar to this prior work, we let the user provide images and texts as input and use a VLM to encode these assets.

\textbf{Sketch-based methods.}
Sketch as an input constraint has been mostly used in the GUI design literature, with sketches containing interface components similar to wireframes. 
\citet{jain2019sketch2code} use a ResNet-based object detection model to convert sketches to JSON objects in real time. Similarly, \citet{mohian2020doodle2app, baule2021automatic, liang2023sketch2wireframe} use different architectures to solve this task as real-time object detection. \citet{ferreira2021automatically} generate synthetic sketches using some heuristics and use them to pre-train the final model. 
Compared to these methods, our approach allows end-to-end generation with user-provided assets, tackling the broad graphical layout generation while previous methods focused on GUIs only.

\textbf{Multimodal Transformer Models.}
Recent work has tried to apply large language models to the layout generation problem. \citet{tang2023layoutnuwa} treats layout generation as a code generation problem, by converting layouts into SVG strings and using CodeLLaMA to solve the problem. \citet{lin2024layoutprompter} use a few-shot prompted GPT, dynamically selecting the in-context examples to be included in the prompt. Additionally, other works \cite{seol2024posterllama,yang2024posterllava, zhu2024automatic} incorporate a vision encoder to handle input images and \cite{cheng2024graphic} leverages a large multimodal model.
Similarly to these, we treat the layout generation task as a code generation problem and utilize VLMs to process image and text inputs.

%% file: sec/3_perliminary_experiments.tex
\section{The value of sketches as a guidance method}

\begin{figure}[tbh]
    \centering
    \includegraphics[width=0.75\linewidth]{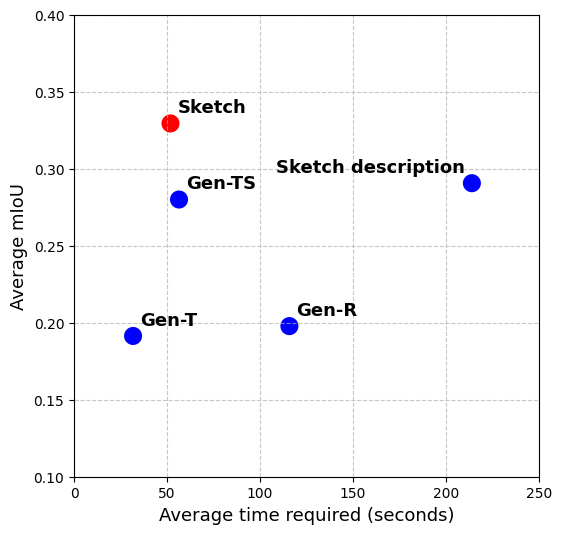}
    \caption{Time-performance trade-off between guidance methods on the PubLayNet dataset.}
    \label{fig:time_guidance}
\end{figure}

We start by assessing empirically the value of user-defined sketches as a guidance method for layout generation.
We use a few-shot ($k$=32) prompted Gemini 1.5 Pro~\cite{gemini15multimodal} and compare the sketch, encoded as an image, to three textual guidance methods from prior work \cite{lin2024layoutprompter}: generation conditioned on asset types (Gen-T), generation conditioned on asset types and sizes (Gen-TS), generation conditioned on spatial relationship between assets (Gen-R). We also compare the efficacy of the sketch to a detailed textual description of the sketch, generated by a captioning model.
Details on few-shot prompt construction and the format for every guidance method are provided in the supplementary material.

We evaluate performance using the maximum Intersection over Union (\textit{mIoU}), i.e. the largest possible IoU 
over all the possible matchings between generated and reference assets. More information about this metric can be found in Sec.~\ref{sec:experimental_setup}. We use the same three datasets as in our experiments. To quantify the time efficiency of each guidance method, we measure the average time required to provide the input. For sketch-based guidance, we measure the time taken to collect each stroke, while for textual constraints, we estimate the time required to write the prompt assuming a typing speed of 200 characters per minute.

The time-performance trade-off of each guidance method measured on our largest dataset, PubLayNet \cite{zhong2019publaynetlargestdatasetdocument}, is shown in Fig.~\ref{fig:time_guidance}. The results clearly demonstrate the superiority of sketches for guiding layout generation, which maximizes performance while at the same time minimizing the time required to form the guidance signal.
We report additional visualizations for the other datasets in in the supplementary material.

%% file: sec/4_methodology.tex
\section{Methodology}

Inspired by previous work in the literature \cite{lin2024layoutprompter,li2024sketch2code}, we formulate the sketch-to-layout problem as a code generation task. Layouts are encoded as protocol buffer strings \cite{protobuf}, with attributes describing the position of assets and their properties. This code representation of the layouts enables a language modeling formulation of the problem. The flexibility of the protocol buffer format allows for straightforward conversion to SVG and therefore image rendering. Outputting a structured representation has several advantages over outputting a layout directly in pixel-space: (i) it can be verified that there are no hallucinations w.r.t.\ the input assets, (ii) the output can be easily interpreted and edited and (iii) it enables interoperability with existing creation tools e.g.\ document editors using a structured representation.

\begin{figure}[tbh]
    \centering
    \includegraphics[width=1\linewidth]{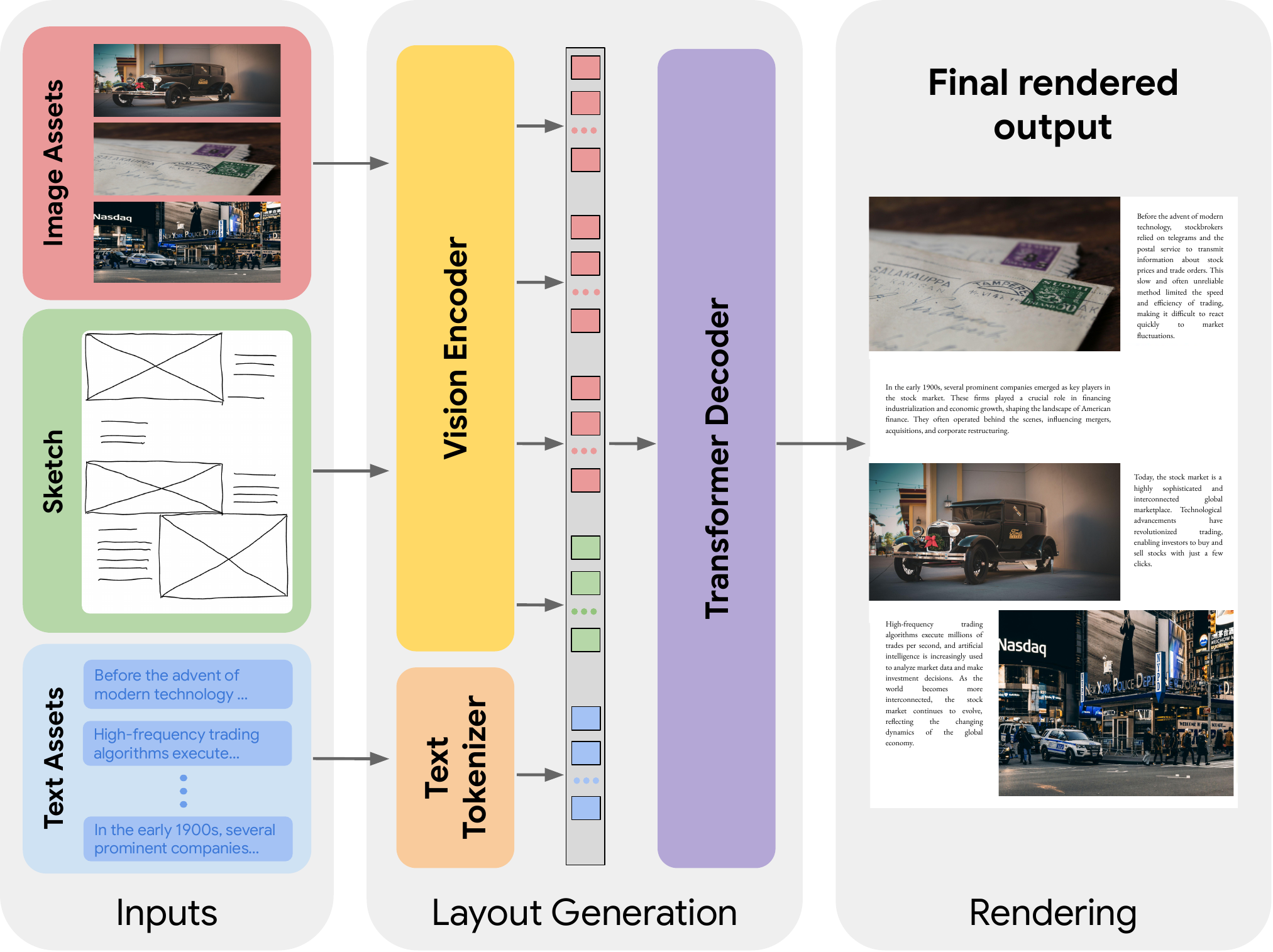}
    \caption{Our method: a sketch, alongside image and text assets are given to a VLM which generates the structured representation format of the layout, which can be rendered as an image.
    }
    \label{fig:pipeline}
\end{figure}

Fine-tuning a VLM to perform well on the novel task of sketch-guided layout generation requires a large amount of human-drawn sketches paired with layouts. Such large-scale data collection is costly and time-consuming. In this section, we first discuss the open-source VLM we adopted to solve the layout generation task and then we introduce a scalable methodology to generate synthetic sketches for model training, while requiring a minimum amount of human data annotation. 

\subsection{Model Structure}

To tackle the problem, we fine-tune PaLIGemma 3B \cite{beyer2024paligemmaversatile3bvlm}, an open-source VLM trained to be versatile and effective to transfer. The language backbone of the architecture consists of Gemma \cite{gemma}, a decoder-only transformer pre-trained on code generation tasks. This makes PaLIGemma well suited for our layout generation task.

The model is multimodal, enabling us to provide both visual and textual inputs to guide the layout generation. An ink-based hand-drawn sketch, outlining the layout structure, is fed into the vision encoder alongside relevant image assets that should appear in the final layout; see also Fig.~\ref{fig:pipeline}. Similarly to previous works on applications of VLMs to videos, the visual backbone of the model is applied independently on each input image, and the patch embeddings are concatenated. Therefore, the ViT \cite{dosovitskiy2020image} serves as a feature extractor, for both the sketch and image assets.
The fact that our model processes image and text content allows the model to understand where to place the assets on canvas, generating a coherent narrative flow.

In addition, a textual prompt specifying the desired layout dimensions, asset names and the content of textual elements, is given as input to the VLM. An example of the prompt can be found in the supplementary material.

\begin{figure}[tbh]
    \centering
    \includegraphics[width=1.0\linewidth]{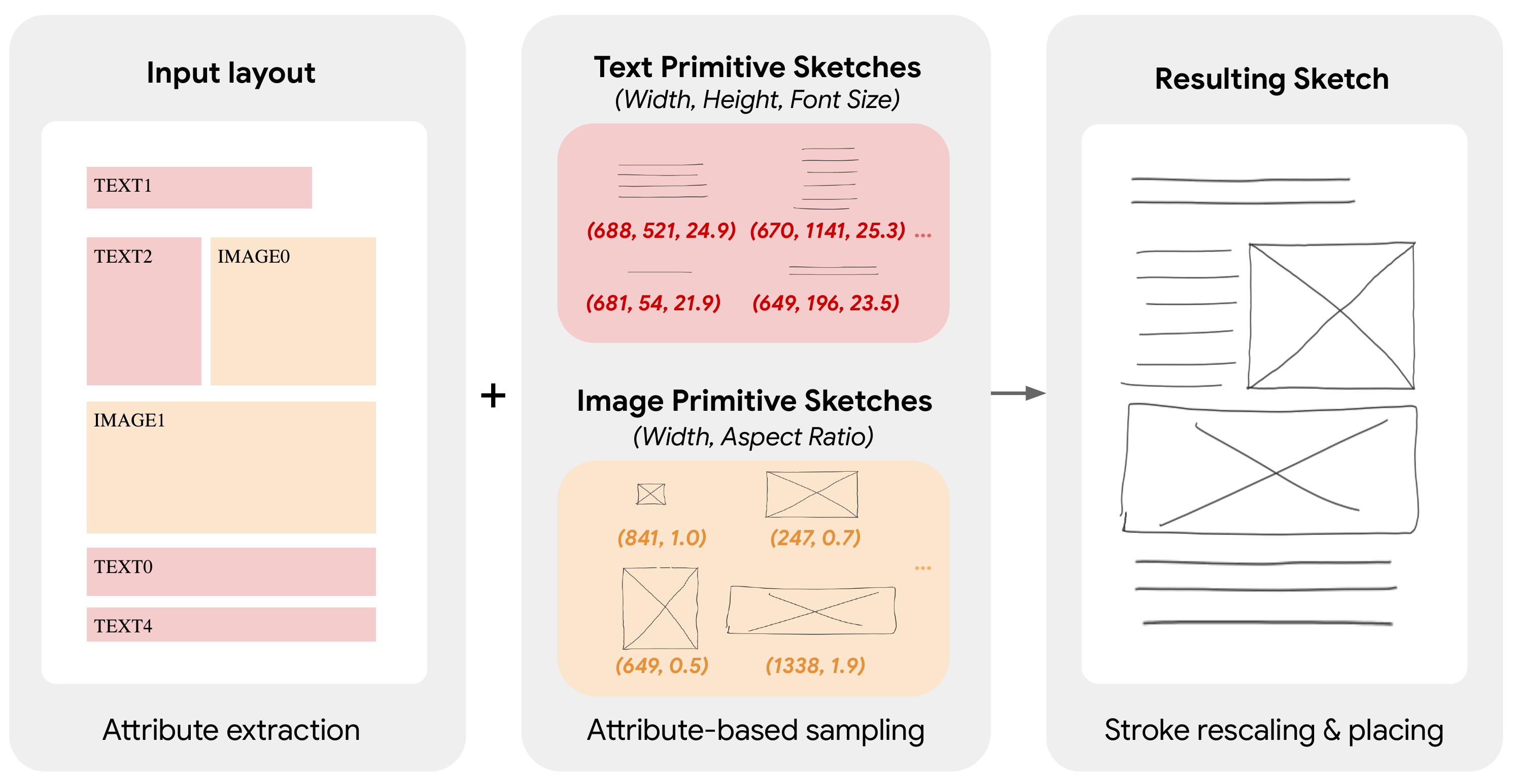}
    \caption{\textbf{Synthetic Sketch Generation Pipeline.} Every asset is matched with a stroke primitive based on its attributes and strokes are rescaled and combined to generate the synthetic sketch.}
    \label{fig:synthetic_sketch}
\end{figure}

\subsection{Synthetic Sketch Generation}

Training the model on sketch-to-layout tasks requires paired data in the form of handwritten sketches and graphic layouts. To the best of our knowledge, there are no publicly available datasets with handwritten sketches resembling the layouts' structure.  However there are document \cite{zhong2019publaynetlargestdatasetdocument, PfitzmannDoclaynet2022} and slide \cite{tanaka2023slidevqadatasetdocumentvisual} datasets, which are particular forms of layouts, without sketches. As direct human sketch annotation at the scale of these datasets is prohibitively costly, we introduce a scalable way to compose a relatively small number of human annotated sketches of layout elements into whole layout synthetic sketches. The methodology is in two steps.

\textbf{Primitive collection.} First, we collected a set of handwritten primitives for image and text assets. Inspired by wireframing, we defined primitives for image and text elements, using one or more horizontal lines to represent a block of text, and a crossed-out rectangle to represent an image. We sampled a set of text and image assets and asked 10 human annotators to draw ink-based sketch primitives on top of these assets, using tablet devices and a custom data collection app. In total, we collect 237 primitives for training and 236 for validation. Since our data consists of image and text assets, we distinguish these two types of primitives. However, the exact same methodology could be extended to an arbitrary number of primitive types for more complex datasets.

\textbf{Synthetic Sketch Composition.} For a given layout, we compose a synthetic sketch by combining sketch primitives for every asset in the layout. More specifically, for each asset, we select a set of $k$ candidate primitives that are the closest in terms of euclidean distance computed on the standardized width and aspect ratio for images and bounding box width, height and font size for texts. Then, from the $k$ closest primitives, we select one at random. The process is conceptually illustrated in Fig.~\ref{fig:synthetic_sketch}.

This methodology doesn't require costly human sketching of full layouts and the annotation time scales linearly with the number of primitives, rather than the number of samples in the training set.
It took $50$ minutes to collect all the primitives necessary to construct our datasets. Meanwhile, it took on average $51.85$ seconds to collect a full sketch for a PubLayNet sample for our test set, $36.54$ seconds for DocLayNet and $13.56$ seconds for SlideVQA. Assuming the same average time required for training and validation sets, collecting sketch data for our datasets would have required $2336$, $292$ and $63$ human hours for PubLayNet, DocLayNet and SlideVQA respectively. More details on the implementation in  Sec.~\ref{synthetic-generation-appendix}.

By using this novel procedure, we obtain a large dataset of layouts paired with sketches: 175k from PubLayNet, 33k from DocLayNet and 27k from SlideVQA. More dataset information and implementation details are provided in the appendix.

%% file: sec/5_experiments.tex
\section{Experiments}
\subsection{Experiment Setup}\label{sec:experimental_setup}

\paragraph{Datasets.}

\begin{table}[tbh]
    \centering 
    \resizebox{\columnwidth}{!}{
    \begin{tabular}{|c|c|c|c|} \hline Dataset Name & \# Training Set & \# Validation Set & \# Human-Collected Test Set \\
    \hline PubLayNet & 162 192 & 900 & 251 \\ 
    DocLayNet & 28 780 & 900 & 268 \\ 
    SlideVQA & 16 593 & 900 & 249 \\
    \hline 
    \end{tabular}
    }
    \caption{Dataset Statistics. While train and validation splits contain synthetically created sketches, the test sets consist of human collected sketches.} \label{dataset_statistics} 
\end{table}

We conduct experiments on three publicly available datasets: PubLayNet \cite{zhong2019publaynetlargestdatasetdocument}, DocLayNet \cite{PfitzmannDoclaynet2022} and SlideVQA \cite{tanaka2023slidevqadatasetdocumentvisual}. Dataset statistics are summarized in Table \ref{dataset_statistics}. 
We carry out training and hyper-parameter tuning on data paired with synthetic sketches, generated as described previously. Training details are provided in \ref{supmat:training_details}.
To fairly assess model performance under real-world conditions, we also collect real human-annotated sketches which we use as a test set.

We train separate models on each dataset and compare their performance to established baselines (detailed in the following section). While training a single model across all datasets could leverage cross-dataset knowledge and potentially improve performance, using separate models for each dataset ensures a fair comparison with baselines, which were exposed to individual datasets only.

\paragraph{Baselines.}
Since the sketch-to-layout task is a novel task and currently under-explored, there is no `ideal baseline` designed to tackle exactly this problem. For this reason, we compare our method against a set of closely related baselines, in order to assess and put our model's performance in perspective with alternative solutions.

We compare our method to LayoutPrompter~\cite{lin2024layoutprompter}, a recent method for conditioned layout generation. LayoutPrompter handles different types of constraints to generate layouts through few-shot prompting. Since \textit{text-davinci-003}, the LLM used by LayoutPrompter, is now deprecated, we substitute it with a few-shot prompted Gemini 1.5 Pro~\cite{gemini15multimodal}, making this baseline even stronger.

LayoutPrompter use the following guidance methods: generation conditioned on asset types (Gen-T), generation conditioned on asset types and sizes (Gen-TS), generation conditioned on spatial relationship between assets (Gen-R). These constraints provide different levels of layout information. More details on these guidance methods and few-shot prompt examples are provided in \ref{supmat:gemini-fewshot-prompts}.

In order to show the value of our synthetic data, we also compare our approach using a fine-tuned small model to a sketch-guided state-of-the-art VLM, Gemini 1.5 Pro with few-shot prompting.

An important difference between our method and LayoutPrompter is that our method is content-aware: it takes as input the sketch \textit{and} the text and image assets. On the contrary, LayoutPrompter's inputs consists of only layout constraints. In order to provide a fair comparison with the baselines, and analyse the effect of providing the content of the assets, we report results in both the content-agnostic and content-aware settings. For the no-content setting, we train our model without providing images and text assets. For the content-aware setup, we add asset content to the few-shot examples for both LayoutPrompter and few-shot Gemini baselines.

\paragraph{Metrics.}
To evaluate model and baseline performance, we use metrics widely adopted in the literature. 

\textit{Intersection over Union (IoU)} and \textit{Maximum Intersection Over Union (mIoU)} \cite{Kikuchi_2021}.
Differently from IoU, where every generated asset is matched to the target asset sharing the same name or identifier, mIoU corresponds to the maximum intersection over union over all the possible matchings between generated and target assets, where elements are paired only depending on their position.

IoU and mIoU are usually limited when evaluating unconstrained layout generation as they reward the model for generating a layout resembling the target, which might not be the only correct way to generate a visually appealing layout with the given assets.
However, in our approach, the user explicitly guides the model towards the target layout using a sketch, specifying where the assets should be positioned. This guidance makes IoU an appropriate metric to measure model performance.

\begin{table*}[tbh]
\renewcommand{\arraystretch}{1.3}
\begin{small}
\resizebox{1\textwidth}{!}{
\begin{tabular}{lccccccccccccccc}
    \toprule
    &   \multicolumn{5}{c}{PubLayNet}  &  \multicolumn{5}{c}{DocLayNet}  &  \multicolumn{5}{c}{SlidesVQA} \\ 
    \cmidrule(l){2-6}  \cmidrule(l){7-11} \cmidrule(l){12-16}
    Method   & IoU $\uparrow$ & COS $\uparrow$ & mIoU  $\uparrow$  & Align.  $\downarrow$  & Overlap  $\downarrow$ & IoU $\uparrow$ & COS $\uparrow$ & mIoU $\uparrow$  & Align.  $\downarrow$  & Overlap  $\downarrow$ & IoU $\uparrow$ & COS $\uparrow$ & mIoU  $\uparrow$  & Align.  $\downarrow$  & Overlap  $\downarrow$  \\ \hline
    \multirow{1}{*}{ LayoutPrompter(Gen-T) w/ content } & 0.13  & 0.52  & 0.21  & \textbf{0.04}  & 0.18  & 0.13  & 0.55  & 0.19  & 1.75  & \textbf{0.03}  & 0.39  & 0.68  & 0.43  & 1.77  & 2.47 \\
    \multirow{1}{*}{ LayoutPrompter(Gen-TS) w/ content } & 0.14  & 0.27  & 0.29  & 0.29  & 0.08  & 0.14  & 0.38  & 0.22  & 2.52  & 0.09  & 0.40  & 0.57  & 0.44  & 6.92  & 2.49 \\
    \multirow{1}{*}{ LayoutPrompter(Gen-R) w/ content } & 0.11  & 0.49  & 0.22  & 0.25  & 0.15  & 0.12  & 0.56  & 0.19  & \textbf{0.79}  & 0.11  & 0.35  & 0.68  & 0.39  & \textbf{0.98}  & \textbf{2.35} \\
    \multirow{1}{*}{ Sketch-guided Gemini w/ content } & 0.15  & 0.33  & 0.32  & 0.31  & 0.08  & 0.15  & 0.42  & 0.25  & 0.93  & 0.05  & 0.40  & 0.63  & 0.46  & 1.85  & 2.43 \\
    \multirow{1}{*}{ \textbf{FT-PaliGemma w/ content (Ours)} } & \textbf{0.62}  & \textbf{0.69}  & \textbf{0.76}  & 0.34  & \textbf{0.03}  & \textbf{0.46}  & \textbf{0.68}  & \textbf{0.59}  & 2.92  & \textbf{0.03}  & \textbf{0.66}  & \textbf{0.79}  & \textbf{0.75}  & 6.54  & 2.42 \\
    \bottomrule 
  \end{tabular}
}
\end{small}
\vspace{-3mm}
\caption{Comparison between Content-Aware FT-PaliGemma and content-aware baselines. $\uparrow$ indicates larger values are better, $\downarrow$ indicates smaller values are better. Alignment values are multiplied by 1000, while Overlap results are multiplied by 10.}
\vspace{-2mm}
\label{Tab:content_aware_metrics_paligemma}
\end{table*}

\begin{table}[tbh]
\renewcommand{\arraystretch}{1.3}
\begin{small}
\resizebox{1.0\linewidth}{!}{
\begin{tabular}{lcccccc}
    \toprule
    &   \multicolumn{2}{c}{DocLayNet}  &  \multicolumn{2}{c}{PubLayNet}  &  \multicolumn{2}{c}{SlideVQA} \\ 
    \cmidrule(l){2-3}  \cmidrule(l){4-5} \cmidrule(l){6-7}
    Method   & IoU$\uparrow$ & COS$\uparrow$ & IoU$\uparrow$ & COS$\uparrow$ & IoU$\uparrow$ & COS$\uparrow$  \\ \hline
    \multirow{1}{*}{Synthetic} & 0.47  & 0.67  & 0.68  & 0.74  & 0.64  & 0.75 \\
    \multirow{1}{*}{ Human } & 0.46  & 0.66  & 0.62  & 0.70  & 0.66  & 0.79   \\
    \bottomrule 
  \end{tabular}
}
\end{small}
\vspace{-3mm}
\caption{Performance comparison of our model on the test set on human sketches vs. synthetic sketches.}
\vspace{-2mm}
\label{Tab:human-vs-synthetic}
\end{table}

\textit{Overlap}~\cite{li2020attributeconditionedlayoutganautomatic} measures the percentage of overlap between generated assets. \textit{Alignment}~\cite{li2020attributeconditionedlayoutganautomatic} measures the graphical alignment for the layout. These metrics are commonly used in the literature. Note, however, that lower values for these metrics are not necessarily better in case of constrained layout generation. Alignment may not be always in agreement with the user intent as depicted by the sketch.

As we claim our approach is content-aware, it is necessary to introduce metrics measuring this awareness, rather than only focusing on the geometric structure of the layout. 

\textit{Content Ordering Score}. Inspired by \cite{li2020attributeconditionedlayoutganautomatic}, we introduce a new metric leveraging the Levenshtein Distance~\cite{levenshteindistance} to measure if the ground truth reading order and narrative flow are preserved in the generated document, taking values between 0 and 1.
To compute the Levenshtein Distance, we take the center of each asset's bounding box, and sort them first by Y-coordinate, then by X-coordinate. This aligns with the intuition of reading in left-to-right orientation languages: assets are sorted top-to-bottom and left-to-right. Then, for a set of asset names $\{ a_k \}_{k=1}^n$, sorted as described above by their center coordinates, we map every asset $a_k$ to a string character $c(a_k)$ and create a sequence $y = concat( c(a_1), \dots, c(a_n) )$. The Content Ordering Score (COS) is computed as
$$ COS = 1 - \frac{\text{lev}(\hat{y}, y)}{\max( |\hat{y}|, |y|)},$$ where $\hat{y}$ is the layout generated by the model, $y$ is the ground truth layout and $\text{lev}(\cdot)$ is the edit distance between assets in two layouts.

\subsection{Main Results}

In what follows, we present the main results of this work. We compare our method against prior techniques, providing results highlighting the effectiveness of our proposed approach. We also provide some ablations studies which deepen our understanding of the method's behaviour.

\begin{figure*}[tbh]
  \centering
  \includegraphics[width=1\linewidth]{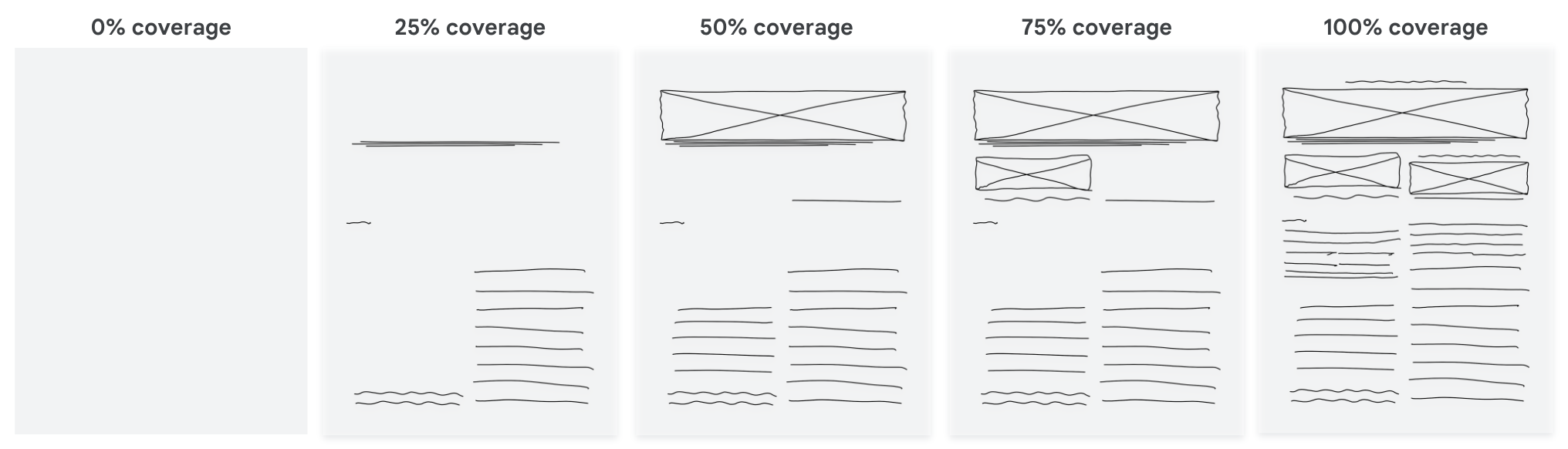}
  \caption{Different coverage rates.}
  \label{fig:coverages}
\end{figure*}

\subsubsection{Content-Aware Solution}

For the content-aware layout generation setting, we compare our approach to Gemini-based LayoutPrompter (Gen-T, Gen-TS and Gen-R) and sketch-guided few-shot prompted Gemini. Results are presented in the Table \ref{Tab:content_aware_metrics_paligemma}.

Our model significantly surpasses the alternative approaches in terms of Maximum IoU, almost achieving a 50\% improvement. This result, alongside qualitative results in the supplementary material, demonstrates our method's ability to correctly place elements within the canvas, carefully following the user-provided sketch structure.

When evaluating our model on synthetic and human-produced sketches, we find comparable performance, demonstrating a minimal distribution shift between our synthetic sketches and full human-produced sketches. We report the results in Table~\ref{Tab:human-vs-synthetic}. This result validates the use of synthetic sketches as proper training data for fine-tuning VLMs on the sketch-to-layout problem.

\begin{table*}[tbh]
\renewcommand{\arraystretch}{1.3}
\begin{small}
\resizebox{1\textwidth}{!}{
\begin{tabular}{lccccccccc}
    \toprule
    &   \multicolumn{3}{c}{PubLayNet}  &  \multicolumn{3}{c}{DocLayNet}  &  \multicolumn{3}{c}{SlidesVQA} \\ 
    \cmidrule(l){2-4}  \cmidrule(l){5-7} \cmidrule(l){8-10}
    Method   & mIoU  $\uparrow$  & Align. $\downarrow$  & Overlap $\downarrow$ & mIoU $\uparrow$  & Align.  $\downarrow$  & Overlap   $\downarrow$ & mIoU $\uparrow$  & Align.  $\downarrow$  & Overlap  $\downarrow$ \\ \hline
    \multirow{1}{*}{ LayoutPrompter(Gen-T) w/o content } & 0.22  & \textbf{0.10}  & 0.12  &  0.18  & 0.48  & 0.08  &  0.39  & \textbf{0.74}  & 2.58  \\
    \multirow{1}{*}{ LayoutPrompter(Gen-TS) w/o content } & 0.33  & 0.37  & 0.08  & 0.24  & 1.95  & 0.08 & 0.43  & 4.28  & 2.40 \\
    \multirow{1}{*}{ LayoutPrompter(Gen-R) w/o content } & 0.23  & 0.36  & 0.19  & 0.18  & 0.79  & 0.12  &  0.36  & 1.45  & 2.49 \\
    \multirow{1}{*}{ Sketch-guided Gemini w/o content } & 0.33  & 0.46  & \textbf{0.02}  & 0.23  & 0.\textbf{23}  & \textbf{0.03}  &  0.47  & 2.76  & \textbf{2.39} \\
    \multirow{1}{*}{ \textbf{FT-PaliGemma w/o content (Ours) } } & \textbf{0.67}  & 0.34  & 0.03  &  \textbf{0.60}  & 2.75  & \textbf{0.03}  & \textbf{0.71}  & 6.73  & 2.44 \\
    \bottomrule 
  \end{tabular}
}

\end{small}
\vspace{-3mm}
\caption{Comparison between Sketch-Only FT-PaliGemma and content-agnostic baselines. $\uparrow$ indicates larger values are better, $\downarrow$ indicates smaller values are better. Alignment values are multiplied by 1000, while Overlap results are multiplied by 10.}
\vspace{-4mm}
\label{Tab:content_unaware_experiment_paligemma}
\end{table*}

\subsubsection{The Importance of Content-Awareness}
To further assess the effectiveness of our content-aware approach, we create a comparative experiment using an approach not leveraging asset content. This allows to directly measure the benefits of incorporating content information. We fine-tune PaliGemma providing the sketch as the only visual input and an auxiliary textual prompt. Differently from before, image and text assets are not provided at this stage, and the textual prompt only contains information about the layout dimensions and asset types. Adopting the same hyper-parameters as before, we fine-tune three separate models, one per dataset. Results are reported in Table~\ref{Tab:content_unaware_experiment_paligemma}.

Despite surpassing baseline performance, the sketch-only method is unable to match the results achieved by the content-aware model. This suggests  that including asset content information boosts performance, as expected. A visual example illustrating the typical improvements obtained through content-awareness can be found in the supplementary material.

\subsection{Ablation Studies}
To rigorously assess the efficacy of our proposed methodology, we conducted a series of ablation studies.

\subsubsection{Partial Sketches}
In our experiments so far we have assumed complete sketches i.e., the sketch covers all assets of the layout. To further assess the model's performance, we introduced scenarios with partial sketches that cover only a subset of layout elements. This allows us to evaluate the model's creative potential when faced with incomplete information.

The way a partial sketch is generated is the following: given a coverage rate $p$ and the set of assets in a layout, each one of them is randomly included in the sketch with probability $p$. We experiments with coverage rates of 0\%, 25\%, 50\%, 75\% and 100\%. An example of partial sketch for different coverage rates is reported in Figure~\ref{fig:coverages}.

\begin{figure*}[tbh]
\centering
\includegraphics[width=0.8\linewidth]{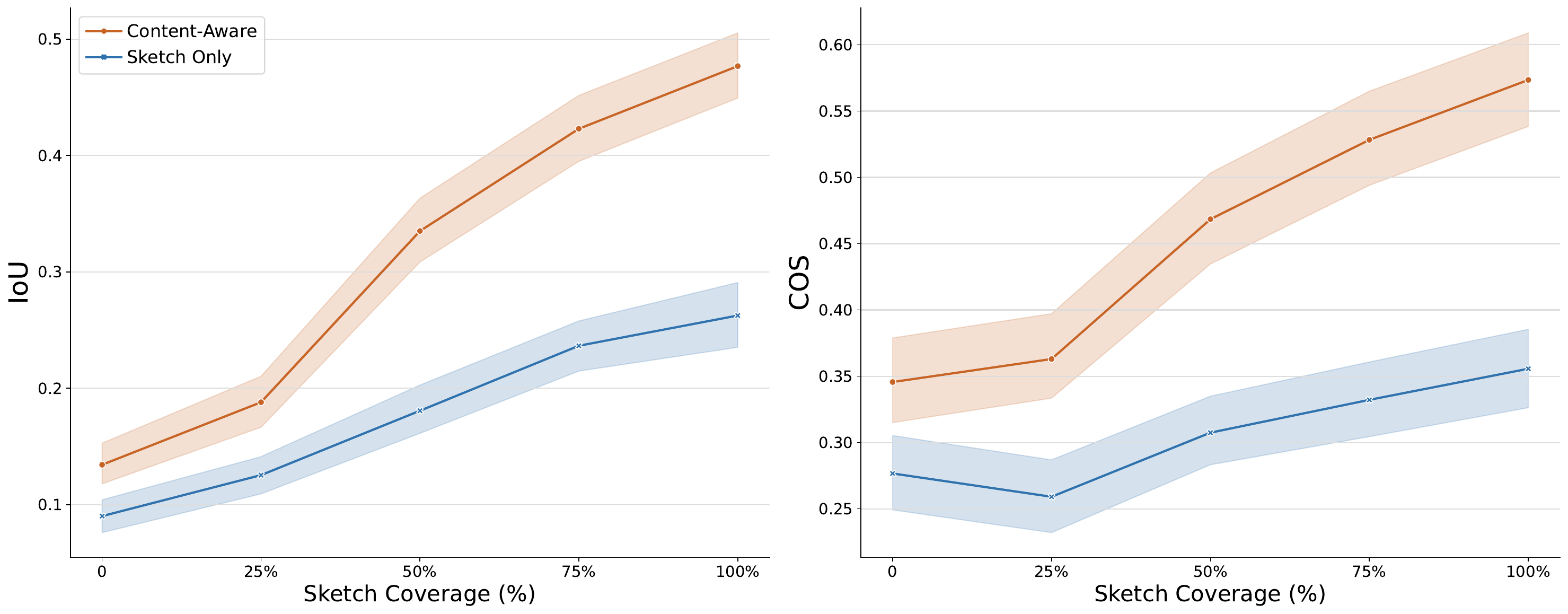}
\caption{Content-aware metrics for different coverage rates of partial sketches, measured on DocLayNet. The blue and the orange line shows the content-agnostic (i.e., sketch-only) and content-aware  comparison correspondingly.}
\label{fig:partial_sketches_metrics}
\end{figure*}

The ablation results are provided in Figure~\ref{fig:partial_sketches_metrics}. We notice a clear trend: increasing sketch coverage correlates with improved IoU, confirming again the sketch's role as a valuable constraint. Similarly, the Content Ordering Score (COS) increases with coverage, as expected due to the sketch's guidance in asset positioning. Notably, both plots show that the content-aware model consistently outperforms its sketch-only counterpart across all coverage levels, confirming our previous findings.

\subsubsection{Content ablations}
\label{sec:shortcuts}

We examine our setting more carefully to identify whether the model exploits shortcuts that can be present in the data. Such shortcuts can help the model place assets without looking at the asset content. We identified and tested a number of settings in which the model can exploit a spurious correlation. 
\begin{itemize}
    \item \textit{Gibberish text}: text asset contents are replaced by random strings containing Latin letters, digits, and whitespaces, aimed at investigating if the model exploit the text sequence length to place text assets.
    \item  \textit{Random images with dimensions}: we replace image pixels with Gaussian noise, keeping the image dimensions in the prompt. This setting potentially eliminates the impact of image dimensions in asset placement. 
    \item \textit{Random images without dimensions}: image pixels are replaced with Gaussian noise, and image dimensions are removed from the prompt. In this setting, no information about the image is given to the model.
    \item  \textit{Full Content}: the original setting where the model has both text and image contents. 
\end{itemize}

\begin{figure}[tbh]
  \centering
  \includegraphics[width=1\linewidth]{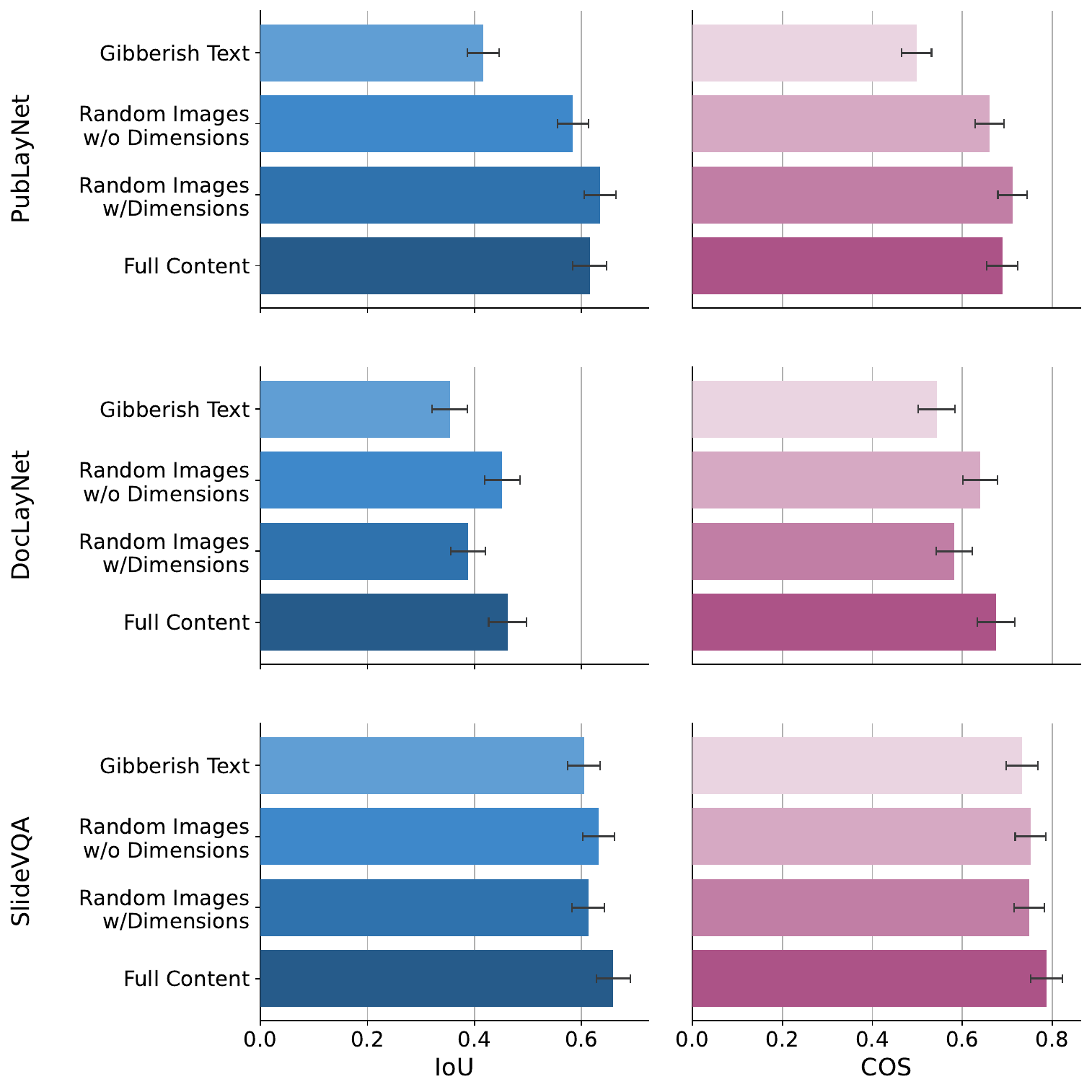}
  \caption{Intersection Over Union and Content Order Score on different settings ablating content.}
\label{fig:shortcuts_ablation}
\end{figure}

Results are shown on Figure~\ref{fig:shortcuts_ablation}. On PubLayNet and DocLayNet, the gibberish text setting performs worse than the original setting in terms of both IoU and COS, suggesting that model does not use text length as a shortcut. On random image settings, we see that removing images does not necessarily lead to a drop in performance. This limitation of our approach can be partially explained by the fact that in our datasets the majority of examples have one image, and only a small part has two or three images, and placing one image given a sketch is a trivial task. Moreover, the PaLIGemma~\cite{beyer2024paligemmaversatile3bvlm} model has not been pretrained on multiple \textit{uncorrelated} images, and achieving understanding of multiple images through a short fine-tuning on a narrow-domain dataset is a difficult task. 

%% file: sec/6_conclusion.tex
\section{Conclusion}

In this work, we present an end-to-end sketch-guided approach for layout generation leveraging VLMs. We motivate the choice of sketches as a guidance method, inspired by how UX designers work, through a comparative analysis with other guidance methods. To train our model, we introduce a novel technique to create synthetic sketches that requires only a few hours of human annotation work and can be scaled to cover large datasets. We release both the human annotated test set and the synthetic train set of sketches. Our fine-tuned model is content-aware and outperforms other constraint-based layout generation methods. Our approach can be generalized to different datasets and domains. More complex sketch primitives can also be added to further guide the model. We encourage future work to apply this methodology to generate sketches for a variety of domains and asset types and train larger, more powerful models to achieve production-level performance.

%% file: sec/7_suppl.tex
\clearpage
\setcounter{page}{1}
\maketitlesupplementary

\section{Implementation Details}

\subsection{Training details}
\label{supmat:training_details}

For all the analysis and experiments described in the paper, the model has been trained for 10 epochs with a batch size of $128$, freezing the ViT and using a cosine learning rate scheduler~\cite{cosinelrscheduler}. The learning rate has been set to $10^{-4}$, and no dropout is used.

During training, the order in which the assets appear in the input textual prompt and the order in which they are fed to the vision encoder are randomized, therefore not matching how they are listed in the output. This serves the specific purpose that the model should learn how to relate each element to the others based on their (image or textual) content, without exploiting any deterministic rule mapping the elements listed in the input to their position in the output.

\subsection{Data Pre-processing}

We performed several pre-processing steps on the three public datasets used in our experiments. First, we crop the content of each bounding box and use an OCR model to extract text content from it. For SlideVQA, we use a large-hole inpainting model to extract the background as a separate asset after masking all foreground bounding boxes.  This allowed us to obtain the content necessary for our content-aware experiments. Then, using the same OCR model we extract the font size and font color of text elements and perform data smoothing of these outputs as a post-processing step. This allowed us to have more accurate rendering for debugging and demonstration purposes. The extracted font size was also used in the synthetic sketch generation pipeline as a clustering attribute.

\subsection{Synthetic Sketch Generation}\label{synthetic-generation-appendix}

To store collected primitives, we use KD-Trees \cite{maneewongvatana1999analysis} but we achieve similar results qualitatively by sampling from the top 10 closest elements iterating over the full dataset of primitives or by sampling at random from pre-computed centroids using K-Means on the training data. KD-Trees have the advantage to not require pre-computation of centroids and are faster than iterating over the full dataset of primitives (log vs linear complexity).

\section{Comparative analysis of the sketch as a guidance method}\label{comparative-analysis-appendix}

The same experiment done on PubLayNet was performed on DocLayNet and SlideVQA. We report the results below.
\begin{figure}[tbh]
    \centering
    \includegraphics[width=0.75\linewidth]{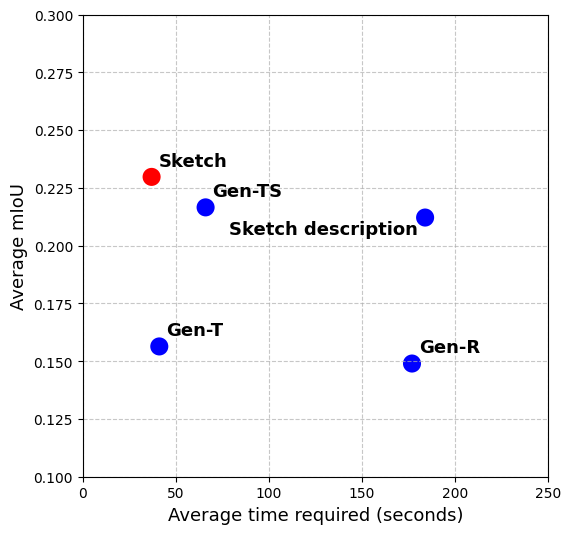}
    \caption{Time-performance trade-off between guidance methods on the DocLayNet dataset.}
    \label{fig:time_guidance}
\end{figure}

\begin{figure}[tbh]
    \centering
    \includegraphics[width=0.75\linewidth]{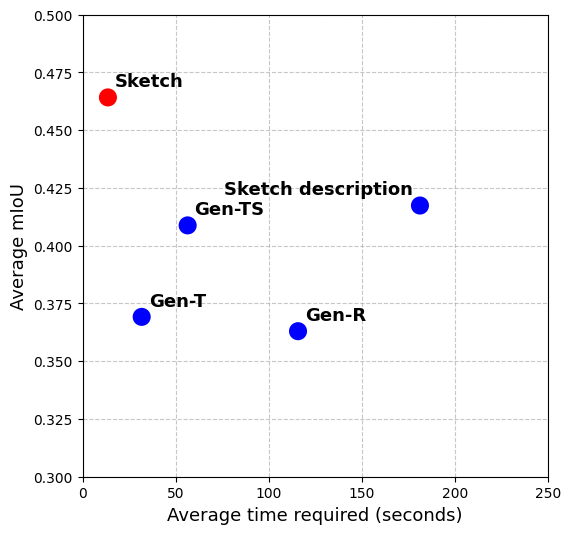}
    \caption{Time-performance trade-off between guidance methods on the SlideVQA dataset.}
    \label{fig:time_guidance}
\end{figure}

\section{Prompt Examples}
For Gen-T, Gen-TS, Gen-R we follow the LayoutPrompter \cite{lin2024layoutprompter} method and use the same prompts, the only difference is that our output format has a JSON structure.

\subsection{Textual Description of the Sketch To Layout}
The task is to generate a layout based on the textual description of the sketch. We determine whether the sketch is an efficient modality by comparing the results of sketch-based generation to the ones conditioned on a textual representation of the same sketch.

\subsubsection{Textual description Creation}
Firstly, to create meaningful representations of the sketch, we few-shot prompt Gemini to generate a textual description of the sketch. The prompt is the following:

\begin{lstlisting}[style=prompt]
You are an AI assistant, you are given a sketch made of ink of a layout that could be a scientific document, a slide or an ad, the sketch represents the position of the assets in the real layout. Images are represented by rectangles with the two diagonals drawn, whereas text (so titles, paragraphs, etc.) is represented by ink lines. 

Please give a detailed and quantitative description of the sketch so that a human could reproduce the layout only based on your textual description. You must be as exhaustive as possible describing all the elements in the sketch. Never mention that an image is represented by a rectangle with two diagonals, directly say it's an image. 

Start with describing the overall structure of the layout such as if it is two column format, Then, describe the positions of the image assets and then text assets one by one using terms like middle, corner, upper, lower, left, right. For each asset, describe the size of the asset with ratios compared to the whole layout. Your description should be very detailed. You will be provided with the exact number of assets, and in particular with the exact number of text assets and image assets.

Input:
The sketch has 8 text assets and 2 image assets.
<<SKETCH>>

Output:
The sketch has 10 assets in total: 8 text assets and 2 image assets.

The sketch is organized on two columns that cover the whole height of the sketch, each column width is half of the whole width of the sketch.

The following is a description from top to bottom, left to right of the whole sketch: there is a small text asset on top of the first column, right below it there is an image asset that occupies 1/4 of the height of the column and the whole width of the first column. Below it there are 3 text assets that occupy the remaining 3/4 of the first column. The second column has on top a small text asset, below there is an image asset that occupies 1/4 of the height of the second column, below there are 3 text assets that cover 3/4 of the second column.

<<OTHER FEW SHOT EXAMPLES>>

Input:
The sketch has N text assets and M image assets.
\end{lstlisting}

\subsubsection{Gemini Few-shot prompt}
\label{supmat:gemini-fewshot-prompts}
After we have obtained few-shot descriptions of sketches for our support samples, we can create the few-shot prompt to query Gemini on Text-to-layout task: 

\begin{lstlisting}[style=prompt]
Please generate a layout based on the given information. You need to ensure that the generated layout looks realistic, with elements well aligned and avoiding unnecessary overlap.

Task Description: generation conditioned on given textual description of the layout

Layout Domain: slide layout

The sketch has 7 assets in total: 5 text assets and 2 image assets.

The sketch represents a slide with an image asset acting as background covering the whole width and height of the slide.

This is a description from top to bottom of the whole sketch. At the top left part of the sketch, there is a text asset, covering 1/4 of the sketch width and 1/4 of the sketch height. Next to it, on its right, there is another text asset, covering 1/4 of the sketch width and 1/3 of the sketch height.

Then, at the bottom left, there is a text asset, covering 1/2 of the sketch width and 1/2 of the sketch height. Next to it, on its right, there is another text asset, covering 1/4 of the sketch width and 1/3 of the sketch height. At the bottom right, there is a text asset, covering 1/4 of the sketch width and 1/8 of the sketch height.

At the bottom left corner, there is an image asset, covering 1/8 of the sketch width and 1/8 of the sketch height. Element Type Constraint: background | image_0 | page_text_0 | page_text_4 | page_text_3 | page_text_1 | other_text_2

{
  "elements": [
    {
      "name": "background",
      "bbox": {
        "width": 1000,
        "height": 1000
      }
    },
    {
      "name": "image_0",
      "bbox": {
        "xmin": 18,
        "ymin": 891,
        "width": 86,
        "height": 91
      }
    },
    {
      "name": "page_text_0",
      "bbox": {
        "xmin": 282,
        "ymin": 92,
        "width": 233,
        "height": 237
      }
    },
    {
      "name": "page_text_4",
      "bbox": {
        "xmin": 471,
        "ymin": 504,
        "width": 286,
        "height": 245
      }
    },
    {
      "name": "page_text_3",
      "bbox": {
        "xmin": 51,
        "ymin": 512,
        "width": 387,
        "height": 258
      }
    },
    {
      "name": "page_text_1",
      "bbox": {
        "xmin": 535,
        "ymin": 94,
        "width": 393,
        "height": 278
      }
    },
    {
      "name": "other_text_2",
      "bbox": {
        "xmin": 732,
        "ymin": 893,
        "width": 242,
        "height": 71
      }
    }
  ]
}

\end{lstlisting}
\subsection{Sketch To Layout Gemini}
To correctly perform few-shot prompting using Gemini, we define two different input formats depending on whether the content has to be included and given as input to the model.

\subsubsection{Sketch-Only to Layout}
To generate the prompt given to the model, we leverage 32 support examples randomly selected each time the model is queried. After providing an initial instruction describing the purpose of the task, we provide a specific set of information for each support sample: the type of layout (slide or document), the description of the primitives used to draw the sketch, the names of the assets appearing in the result, the corresponding sketch and its protobuf representation. The following is an example showing how a DocLayNet sample is leveraged when using it as support:

\begin{lstlisting}[style=prompt]
Please generate a layout based on the given information. You need to ensure that the generated layout looks realistic, with elements well aligned and avoiding unnecessary overlap.

Task Description: generation conditioned on given element types and sketch

Layout Domain: document layout.

To generate the layout you must follow the sketch represented in the next image, where each image asset is represented by a crossed rectangle, whereas text assets (titles, paragraphs, descriptions, ...) are represented by straight or wavy horizontal lines, in particular each cluster of straight horizontal lines (that could contain any number of lines starting from 1) represent one text asset.
Element Type Constraint: picture 0 | picture 1 | picture 2 | text 3 | text 4 | text 5
\end{lstlisting}

The instruction is then followed by the sketch, in image format, and the protobuf representation. As we are working in the sketch-only setting, no information about the assets' content is provided, and only their names are listed. The way the assets are listed and the information are encoded is equivalent to what has been done for the textual baseline, in order to fairly compare the validity of the sketch.
\subsubsection{Sketch with Content to Layout}
Differently from what has been described before, it is now necessary to include the content of each asset in the prompt. Additionally, such a baseline is used to better measure the performance of Content-Aware PaliGemma. Therefore, for a fair comparison, we use the same input format. For each sample used for support, the prompt is as follows:

The text, which contains the content of textual elements given the content-aware nature of the approach, is then followed by the sketch and the output in protobuf format. While image assets for the support samples are not provided in order not to increase the length of the context too much, those belonging to the sample to evaluate are added and appended immediately after the sketch. 
\subsection{Layout Prompter Details}

\begin{lstlisting}[style=prompt]
Please generate a layout based on the given information. You need to ensure that the generated layout looks realistic, with elements well aligned and avoiding unnecessary overlap.

Task Description: generation conditioned on given element types

Layout Domain: slide layout

Canvas Size: canvas width is 160px, canvas height is 120px

Element Type Constraint: background 0 | figure 1 | page_text 2 | title 3

Asset Contents:
background 0: 
<PIL.PngImagePlugin.PngImageFile image mode=RGB size=1024x768 at 0x7111DF0E1310>
figure 1: 
<PIL.PngImagePlugin.PngImageFile image mode=RGB size=1010x607 at 0x7111DF0BBD90>
page\_text 2: Journey Map
title 3: UX LX CONFERENCE JOURNEY
<html>
<body>
<div class="canvas" style="left: 0px; top: 0px; width: 160px; height: 120px"></div>
<div class="background" style="index: 0; left: 0px; top: 0px; width: 160px; height: 120px"></div>
<div class="figure" style="index: 1; left: 2px; top: 13px; width: 157px; height: 94px"></div>
<div class="page\_text" style="index: 2; left: 8px; top: 9px; width: 13px; height: 2px"></div>
<div class="title" style="index: 3; left: 26px; top: 7px; width: 66px; height: 3px"></div>
</body>
</html>
\end{lstlisting}

\subsection{Sketch to Layout Content-Aware PaliGemma}\label{paligemma_textual_prompt}
As explained in the main section, the model is given both textual and image assets information in the input, in order to guide the generation. The following is an example of prompt used when training and evaluating out content-aware solution.

\begin{lstlisting}[style=prompt]
Please prepare a width: 1700 x height: 2200 layout for the following assets: 

text7: Fig. 2 shows the time course changes in normalized rmsEMG of m.MG, m.LG, and m.SOL. The rmsEMG in those muscles increased similarly with increasing exercise intensity. The rmsEMG of m.MG for each of the first 30 s at 20%, 30%, 50%, 60%, 70%, and 80% MVC differed significantly from that during the 30 s of exercise immediately before (i.e., prior intensity) (p < 0.05). Throughout the exercise, the change in rmsEMG of m.MG was largest in the three muscle groups.; 

text5: Fig. 3A shows the time course of changes in intramuscular pH. We found that pH was relatively constant, from resting values (7.06 +/- 0.01) until 60% MVC (7.04 +/- 0.08), but it decreased significantly (p < 0.05) at 70% MVC and with exercise progression, being 6.78 +\- 0.22 at the end of exercise.; 

text3: Fig. 3B shows the time course changes in intramuscular PCr. We found that there were significant differences after the last 30 s at 40% MVC when compared with the value obtained during the first 30 s at 10% MVC (p < 0.05), and that PCr decreased with progression of exercise. Above 70% MVC, the values were significantly different when compared with those obtained during the 30 s of exercise immediately before. A linear regression line was drawn to obtain the highest correlation coefficient above the last 30 s of 40% MVC, at which significant difference was;

text0: Division of data analysis (30s).;

text1: course changes in each parameter, and Fisher's PLSD post hoc comparisons were used to determine the significance of differences of each parameter every 30 s. A linear regression analysis was used to examine the relationship between each parameter. P < 0.05 was defined as statistically significant.;

text2: 02mus measurement (6 s; once per three contraction phases).;

text6: Figure I Procedure for data analysis. Each parameter was analyzed every 30 s. Muscle phosphocreatine (PCr), inorganic phosphate (Pi), pH, estimated ADP and free energy of ATP hydrolysis (AGATP), pulmonary oxygen uptake (V02pul), and electromyogram (EMG) were averaged over 30 s. The data for muscle oxygen consumption (VO2mus) were obtained during the third (20-26 s) and sixth (50-56 s) contractions at each intensity. The V 02mus value of the third contraction was used to represent the first 30 s of each minute, whereas the V 02mus value of the sixth contraction was used to represent the last 30 s of each minute.; 

title4: Results; 

figure0 (width: 1386 x height: 765): <image>. The output should be a single sentence, in protocol buffer debug string format.
\end{lstlisting}

When running our ablation study assessing the usefulness of adding the assets' content to the input, we avoid including text contents and images to the prompt, as the only considered visual input is the sketch. Therefore, only text elements are included, reporting their dimensions but not their content.
\clearpage
\section{Content-Agnostic vs Content-Aware Results}
Incorporating the content of the assets in addition to the sketch helps the model to better place the assets, especially in cases where the positions of the assets are correct but the order of them is incorrect. Such an example can be seen in Figure \ref{fig:content_aware_generation} where the content-agnostic placement was incorrect due to the misorder of the elements.

\begin{minipage}{0.95\textwidth}
    \centering
    \vspace{5mm}
    \includegraphics[width=1.0\textwidth]{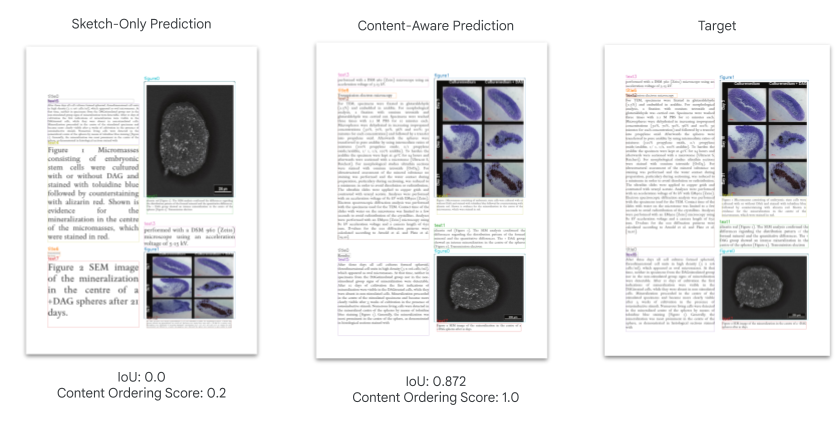}
    \captionof{figure}{Providing additional assets information helps the model better generate the desired layout.\protect\footnotemark[2]}
    \label{fig:content_aware_generation}
\end{minipage}

\clearpage
\section{Complete Partial Sketches Results}
The results for partial sketches ablation study on all the datasets can be seen in Figure \ref{fig:partial_sketches_complete}. It can be observed that increasing the coverage yields better results, confirming the value of sketch as a guidance prior. However, this increase is not monotonic as can be seen in the increase from 75\% to 100\% on PubLayNet and 0\% to 25\% on DocLayNet.
\begin{minipage}{0.95\textwidth}
    \centering
    \vspace{5mm}
    \includegraphics[width=\linewidth]{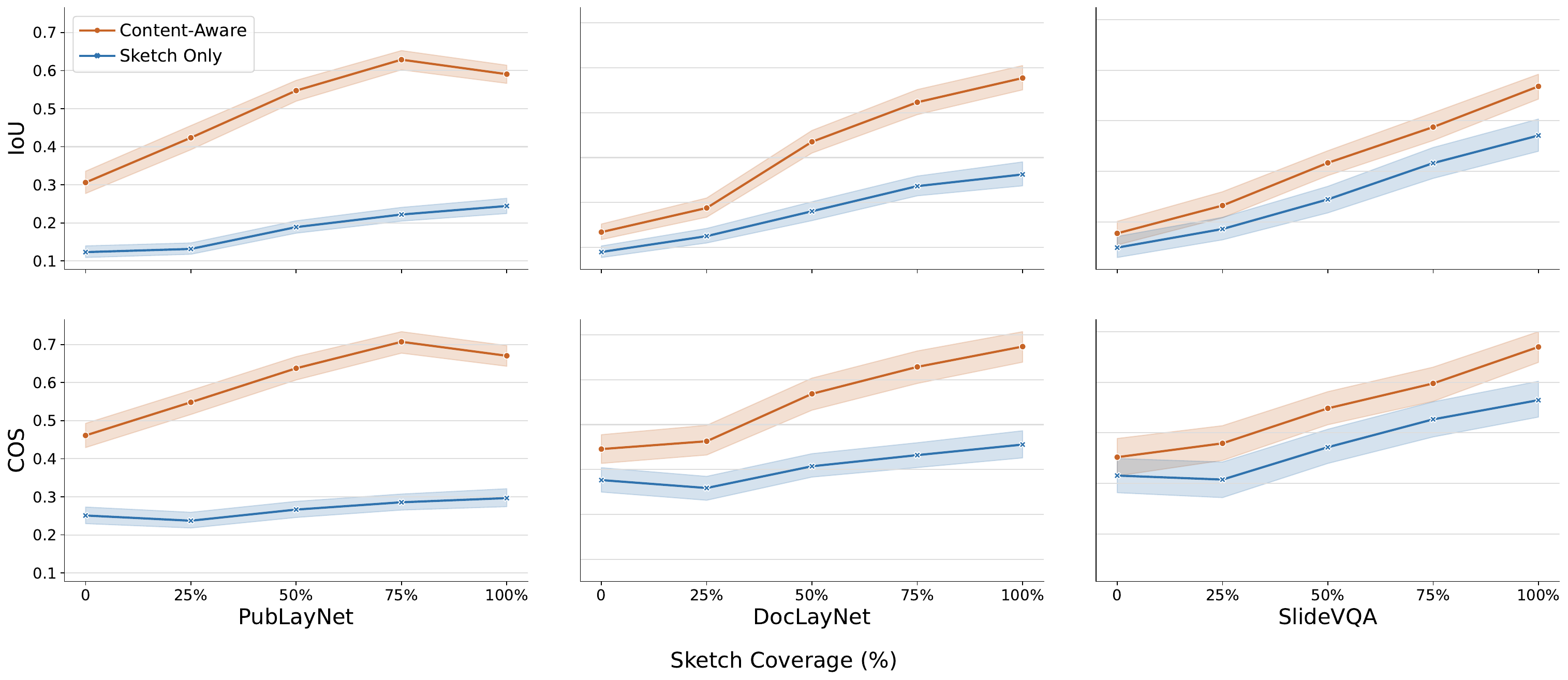}
    \captionof{figure}{Partial sketch results on all datasets.}
    \label{fig:partial_sketches_complete}
\end{minipage}
\section{Synthetic vs Real sketches}\label{synthetic-real}
The complete results on synthetic and real sketches can be seen in the Table \ref{Tab:human_synthetic_comparison} below. The alignment and the overlap metrics of the original layouts are also given in the last two columns, which can be interpreted as reference values that good layouts would have similar values to. There is no statistically significant difference between the metrics for the synthetic and human collected sketches, which confirms that the synthetic sketches are similar to actual sketches. 
   
 \begin{small}
 \vspace{5mm}
 \resizebox{1\textwidth}{!}{
  \begin{tabular}{llrrrrrrr}
    Dataset & Setting  & \multicolumn{1}{c}{mIoU} & \multicolumn{1}{c}{IoU} & \multicolumn{1}{c}{Overlap} & \multicolumn{1}{c}{Alignment} & \multicolumn{1}{c}{COS} & \multicolumn{1}{c}{Alignment Target} & \multicolumn{1}{c}{Overlap Target} \\
    \midrule
    \multirow[c]{2}{*}{DocLayNet} & Human sketches & 0.590 $\pm$ 0.171 & 0.457 $\pm$ 0.252 & 0.003 $\pm$ 0.007 & 0.003 $\pm$ 0.0074 & 0.665 $\pm$ 0.296 & 0.003 $\pm$ 0.008 & 0.0001 $\pm$ 0.001 \\
     & Synthetic sketches & 0.592 $\pm$ 0.164 & 0.466 $\pm$ 0.245 & 0.009 $\pm$ 0.031 & 0.003 $\pm$ 0.007 & 0.669 $\pm$ 0.298 & 0.003 $\pm$ 0.007 & 0.0001 $\pm$ 0.001 \\
    \hline
    \multirow[c]{2}{*}{PubLayNet} & Human sketches & 0.761 $\pm$ 0.132 & 0.623 $\pm$ 0.232 & 0.003 $\pm$ 0.006 & 0.0003 $\pm$ 0.0009 & 0.699 $\pm$ 0.253 & 0.0002 $\pm$ 0.0005 & 0.0004 $\pm$ 0.001 \\
    & Synthetic sketches & 0.806 $\pm$ 0.117 & 0.675 $\pm$ 0.216 & 0.005 $\pm$ 0.010 & 0.0003 $\pm$ 0.001 & 0.741 $\pm$ 0.243 & 0.0002 $\pm$ 0.0005 & 0.0004 $\pm$ 0.001 \\
    \hline
    \multirow[c]{2}{*}{SlideVQA} & Human sketches & 0.747 $\pm$ 0.136 & 0.659 $\pm$ 0.226 & 0.238 $\pm$ 0.136 & 0.006 $\pm$ 0.010 & 0.787 $\pm$ 0.248 & 0.006 $\pm$ 0.010 & 0.236 $\pm$ 0.139 \\
     & Synthetic sketches & 0.752 $\pm$ 0.132 & 0.637 $\pm$ 0.237 & 0.240 $\pm$ 0.134 & 0.008 $\pm$ 0.013 & 0.755 $\pm$ 0.271 & 0.006 $\pm$ 0.010 & 0.235 $\pm$ 0.138 \\
    \bottomrule
    \end{tabular}
 }
\end{small}
\captionof{table}{Comparison between Synthetic and Human Collected Sketches.}
\label{Tab:human_synthetic_comparison}

\FloatBarrier
\clearpage

\begin{figure*}[htbp]
    \centering
    \begin{overpic}[width=0.8\paperwidth]{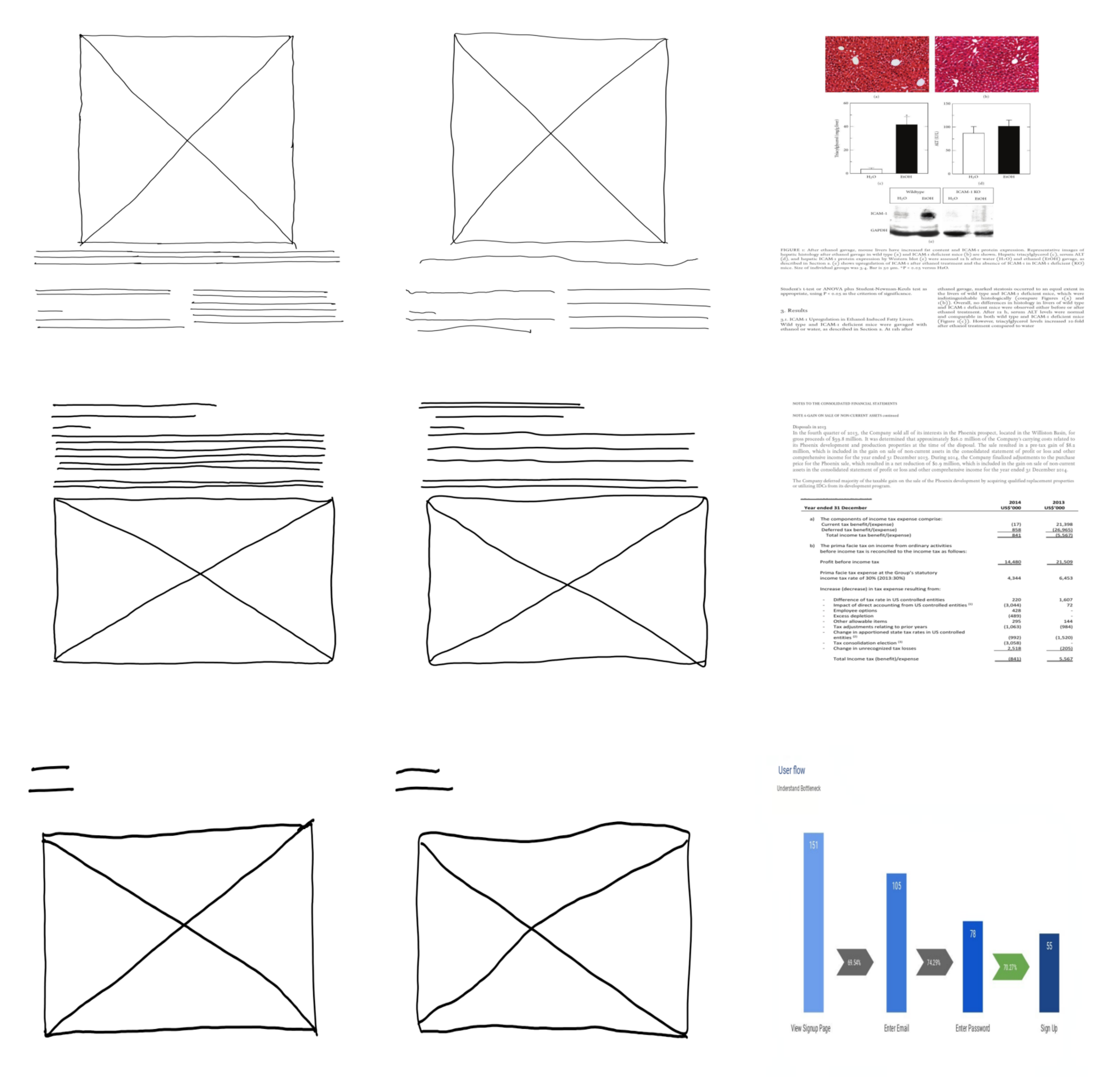}
        \put(9, 99){ Synthetic Sketch}
        \put(44, 99){ Real Sketch}
        \put(80, 99){ Target}

        \put(-2, 74){\rotatebox{90}{ PublayNet Example\footnotemark[3]}}
        \put(-2, 42){\rotatebox{90}{ DoclayNet Example\footnotemark[4]}}
        \put(-2, 10){\rotatebox{90}{ SlidesVQA Example\footnotemark[5]}}
    \end{overpic}
    \caption{Some example layouts with corresponding synthetic and human collected sketches.}
    \label{fig:sketchgrid1}
\end{figure*}

\begin{figure*}[htbp]
    \centering
    \begin{overpic}[width=0.8\paperwidth]{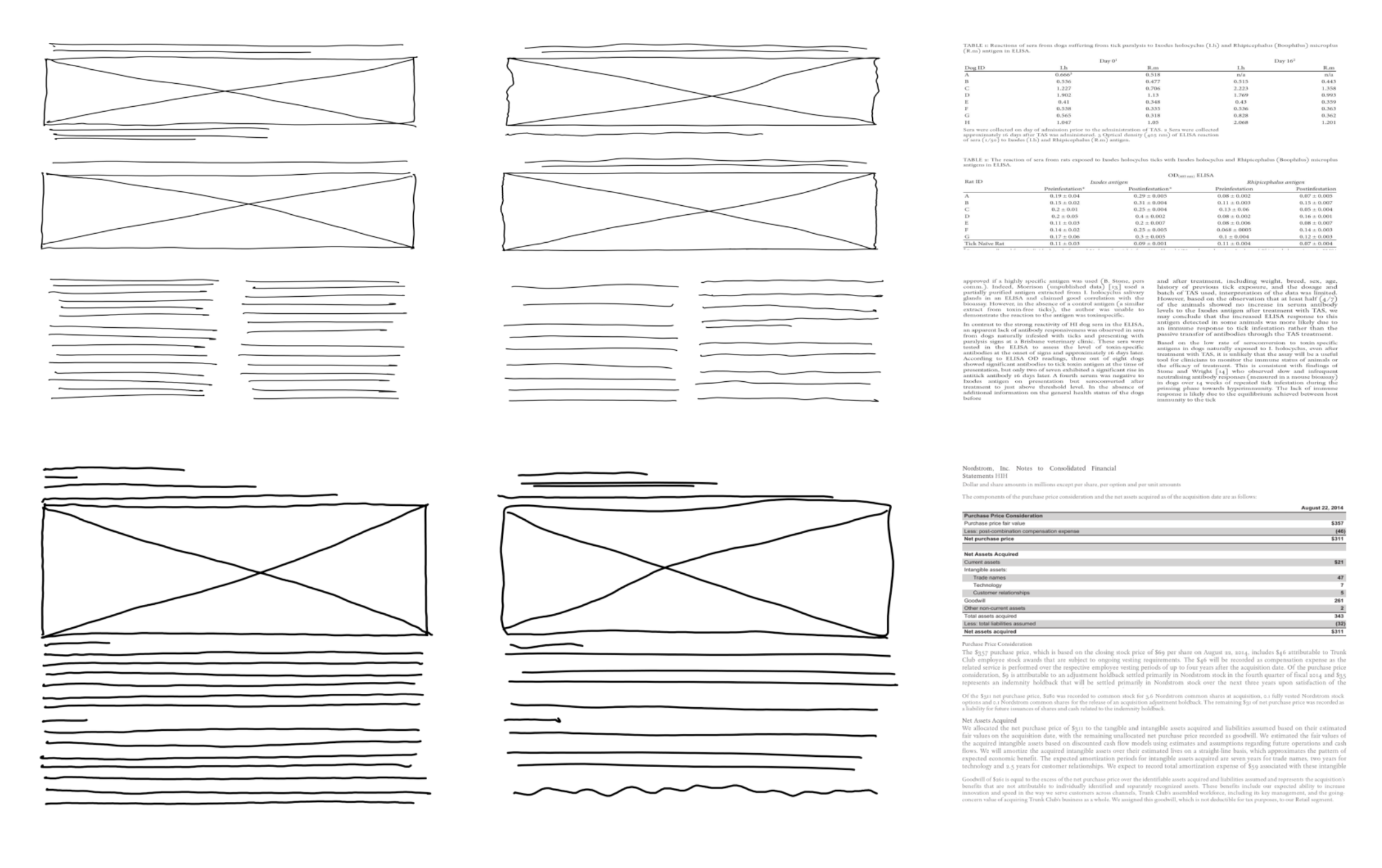}
     \put(9, 65){ Synthetic Sketch}
        \put(44, 64){ Real Sketch}
        \put(80, 64){ Target}

        \put(-2, 37){\rotatebox{90}{ PublayNet Example\footnotemark[6]}}
        \put(-2, 7){\rotatebox{90}{ DoclayNet Example\footnotemark[4]}}
    \end{overpic}
    \caption{More example layouts with corresponding synthetic and human collected sketches.}
    \label{fig:sketchgrid2}
\end{figure*}

\clearpage
\newpage

\section{Qualitative Results}\label{qualitative-results} 
\begin{figure*}[htbp]
    \centering
    \begin{overpic}[width=\textwidth,grid=false,tics=10]{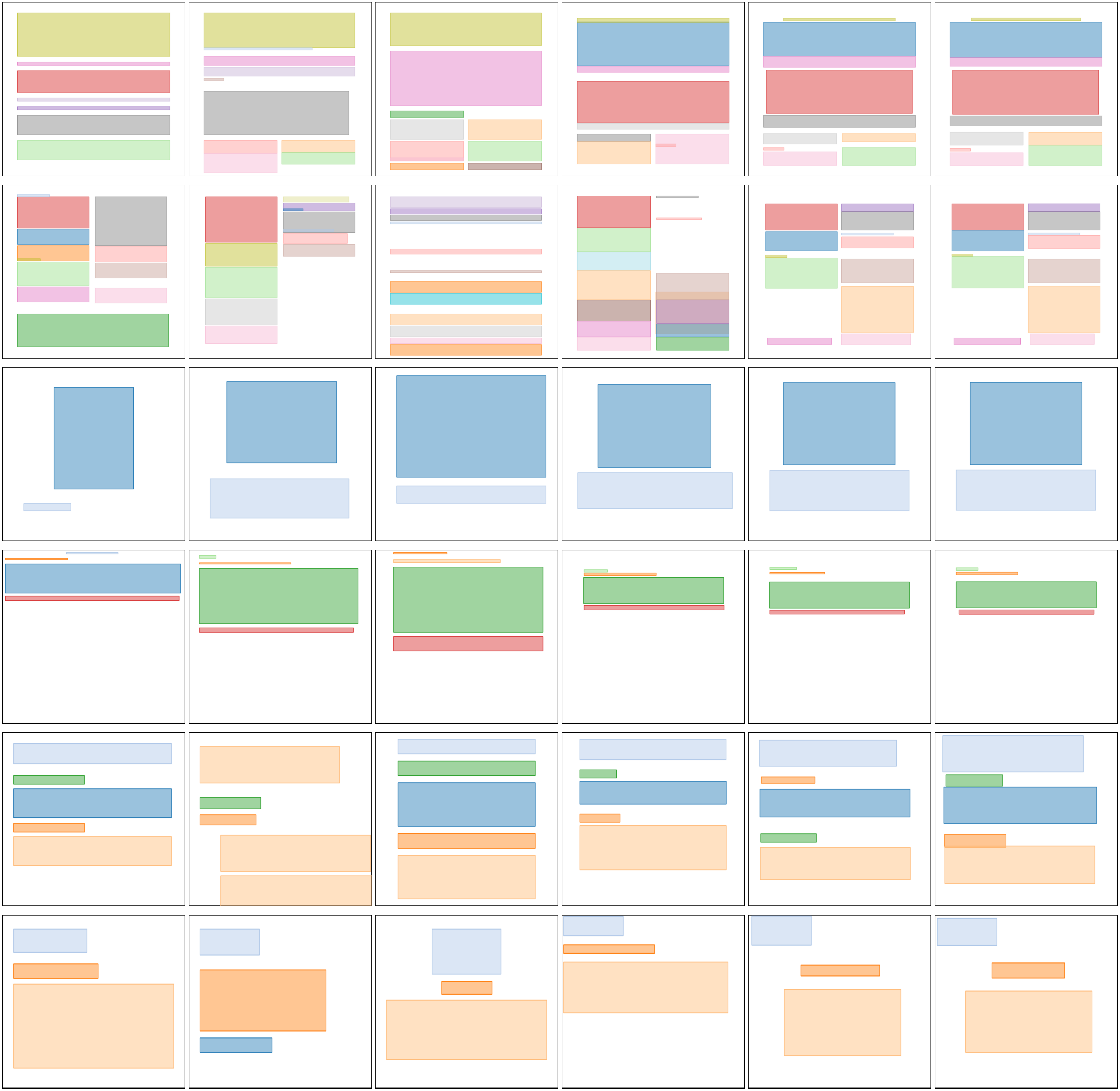}
        \put(2, 99){\scriptsize LayoutPrompter (Gen-T)}
        \put(18, 99){\scriptsize LayoutPrompter (Gen-TS)}
        \put(35, 99){\scriptsize LayoutPrompter (Gen-R)}
        \put(53, 99){\scriptsize Few-shot Gemini}
        \put(71, 99){\scriptsize Sketch2Layout}
        \put(90, 99){\scriptsize Target}
        
        \put(-2, 83.5){\rotatebox{90}{\scriptsize PubLayNet Example}}
        \put(-2, 67.5){\rotatebox{90}{\scriptsize PubLayNet Example}}
        \put(-2, 51){\rotatebox{90}{\scriptsize DocLayNet Example}}
        \put(-2, 35){\rotatebox{90}{\scriptsize DocLayNet Example}}
        \put(-2, 19){\rotatebox{90}{\scriptsize SlidesVQA Example}}
        \put(-2, 2.5){\rotatebox{90}{\scriptsize SlidesVQA Example}}
    \end{overpic}
    \caption{More examples of layouts generated by different methods and our model given the set of assets.}
    \label{fig:grid_2}
\end{figure*}

\begin{figure*}[htbp]
    \centering
    \begin{overpic}[width=\textwidth,grid=false,tics=10]{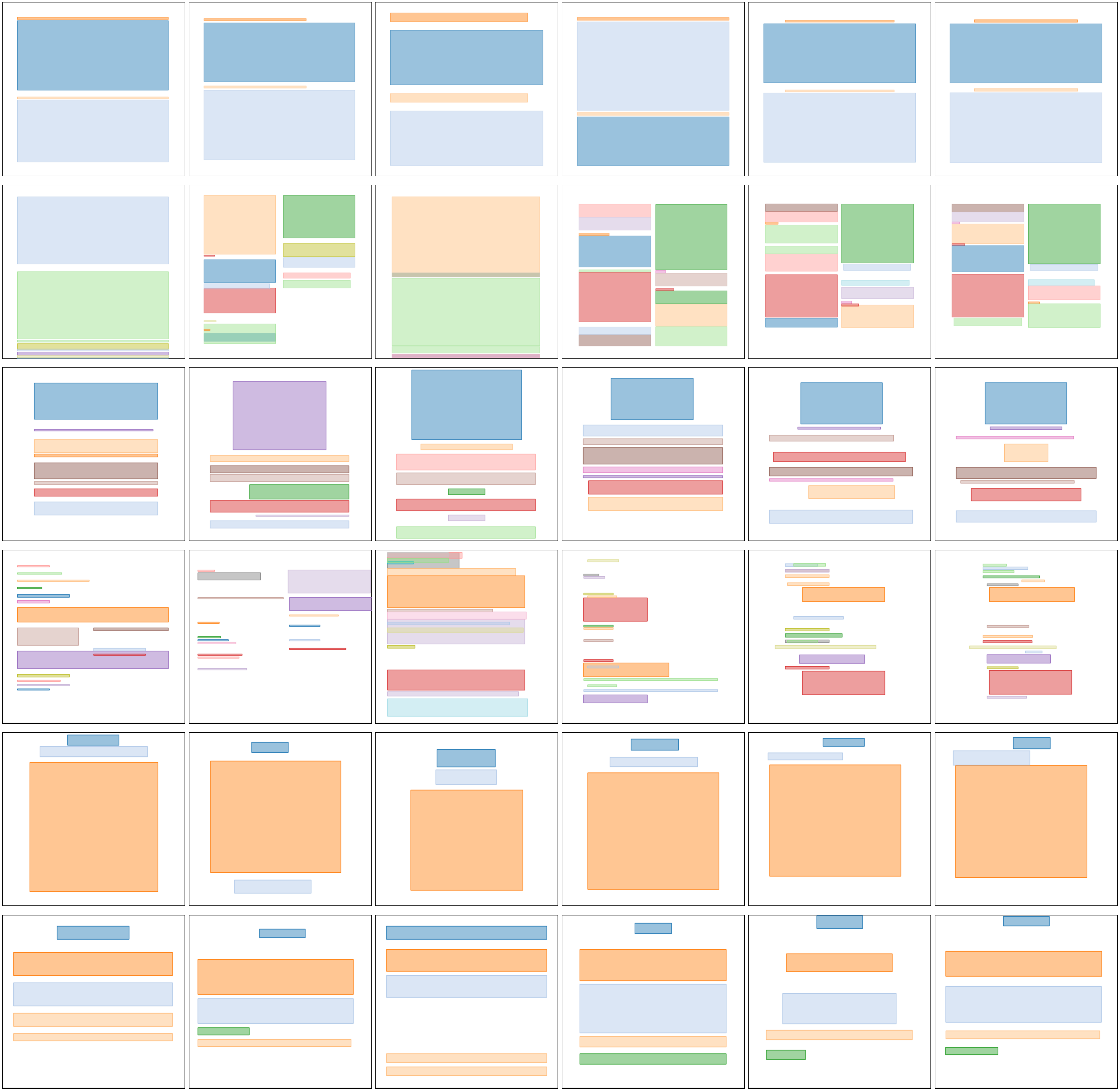}
        \put(2, 99){\scriptsize LayoutPrompter (Gen-T)}
        \put(18, 99){\scriptsize LayoutPrompter (Gen-TS)}
        \put(35, 99){\scriptsize LayoutPrompter (Gen-R)}
        \put(53, 99){\scriptsize Few-shot Gemini}
        \put(71, 99){\scriptsize Sketch2Layout}
        \put(90, 99){\scriptsize Target}
        
        \put(-2, 83.5){\rotatebox{90}{\scriptsize PubLayNet Example}}
        \put(-2, 67.5){\rotatebox{90}{\scriptsize PubLayNet Example}}
        \put(-2, 51){\rotatebox{90}{\scriptsize DocLayNet Example}}
        \put(-2, 35){\rotatebox{90}{\scriptsize DocLayNet Example}}
        \put(-2, 19){\rotatebox{90}{\scriptsize SlidesVQA Example}}
        \put(-2, 2.5){\rotatebox{90}{\scriptsize SlidesVQA Example}}
    \end{overpic}
    \caption{More examples of layouts generated by different methods and our model given the set of assets.}
    \label{fig:grid_3}
\end{figure*}

Qualitative results of our method can be seen on Figure \ref{fig:grid}, \ref{fig:grid_2} and \ref{fig:grid_3} where the assets are shown as boxes with different colors specifying different assets. It can be seen that our method can generate layouts which are more accurate both in terms of the positioning and the ordering of the assets compared to LayoutPrompter(Gen-T, Gen-TS, Gen-R) and few-shot Gemini. 
 
\begin{minipage}{1\textwidth}
\centering
\vspace{7mm}
      \begin{overpic}[width=0.95\textwidth,grid=false,tics=10]{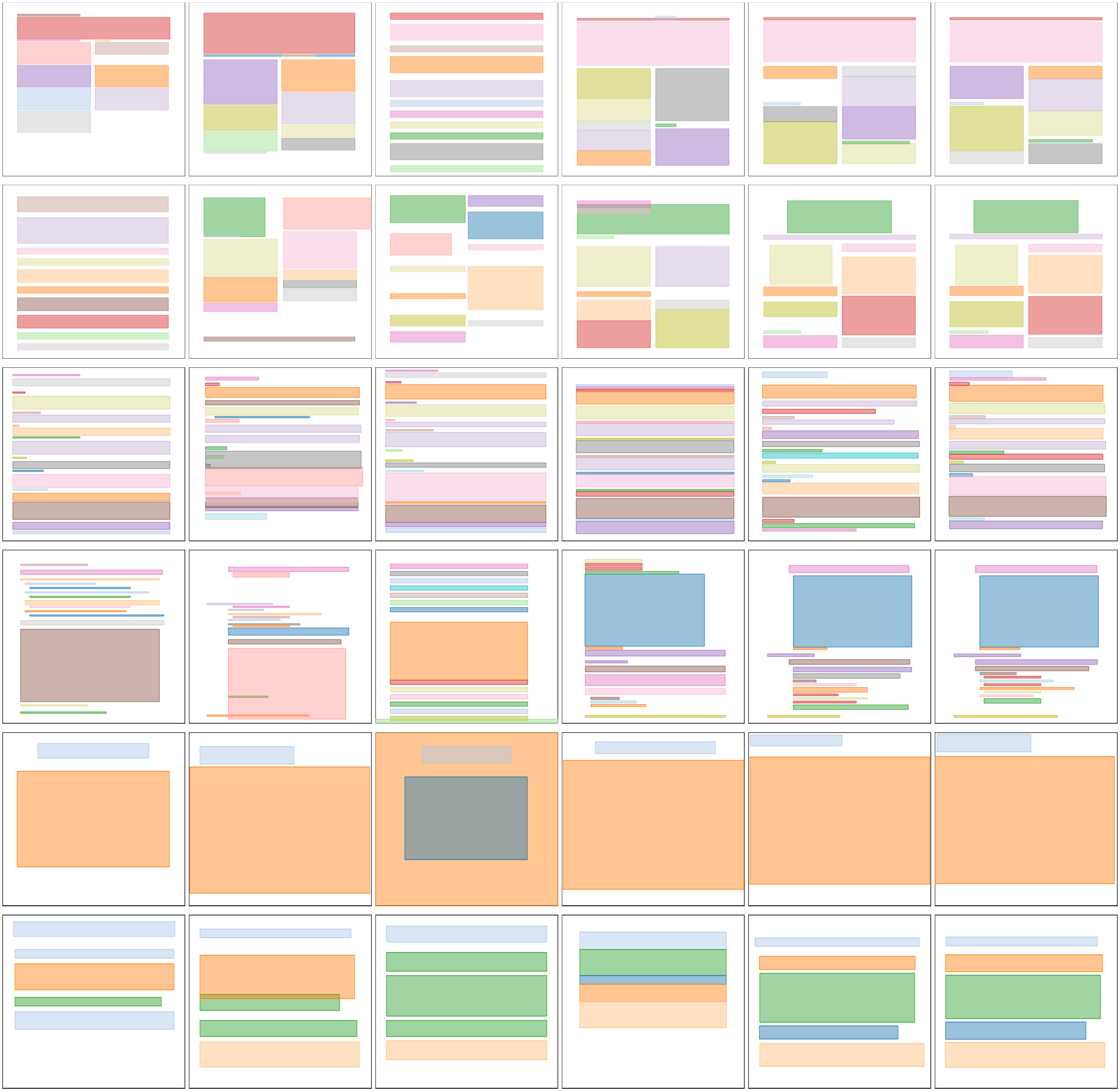}
        \put(2, 99){\scriptsize LayoutPrompter (Gen-T)}
        \put(18, 99){\scriptsize LayoutPrompter (Gen-TS)}
        \put(35, 99){\scriptsize LayoutPrompter (Gen-R)}
        \put(53, 99){\scriptsize Few-shot Gemini}
        \put(71, 99){\scriptsize Sketch-to-Layout}
        \put(90, 99){\scriptsize Target}
        
        \put(-2, 83.5){\rotatebox{90}{\scriptsize PubLayNet Example}}
        \put(-2, 67.5){\rotatebox{90}{\scriptsize PubLayNet Example}}
        \put(-2, 51){\rotatebox{90}{\scriptsize DocLayNet Example}}
        \put(-2, 35){\rotatebox{90}{\scriptsize DocLayNet Example}}
        \put(-2, 19){\rotatebox{90}{\scriptsize SlidesVQA Example}}
        \put(-2, 2.5){\rotatebox{90}{\scriptsize SlidesVQA Example}}
    \end{overpic}
    \captionof{figure}{Examples of layouts generated by different methods and our model given the set of assets. Different assets are identified with different colors, showing the capability of different models to process asset content.}
    \label{fig:grid}
\end{minipage}
\clearpage

\begin{figure*}[htbp]
    \centering
    \begin{overpic}[height=1\textheight,grid=false,tics=10]{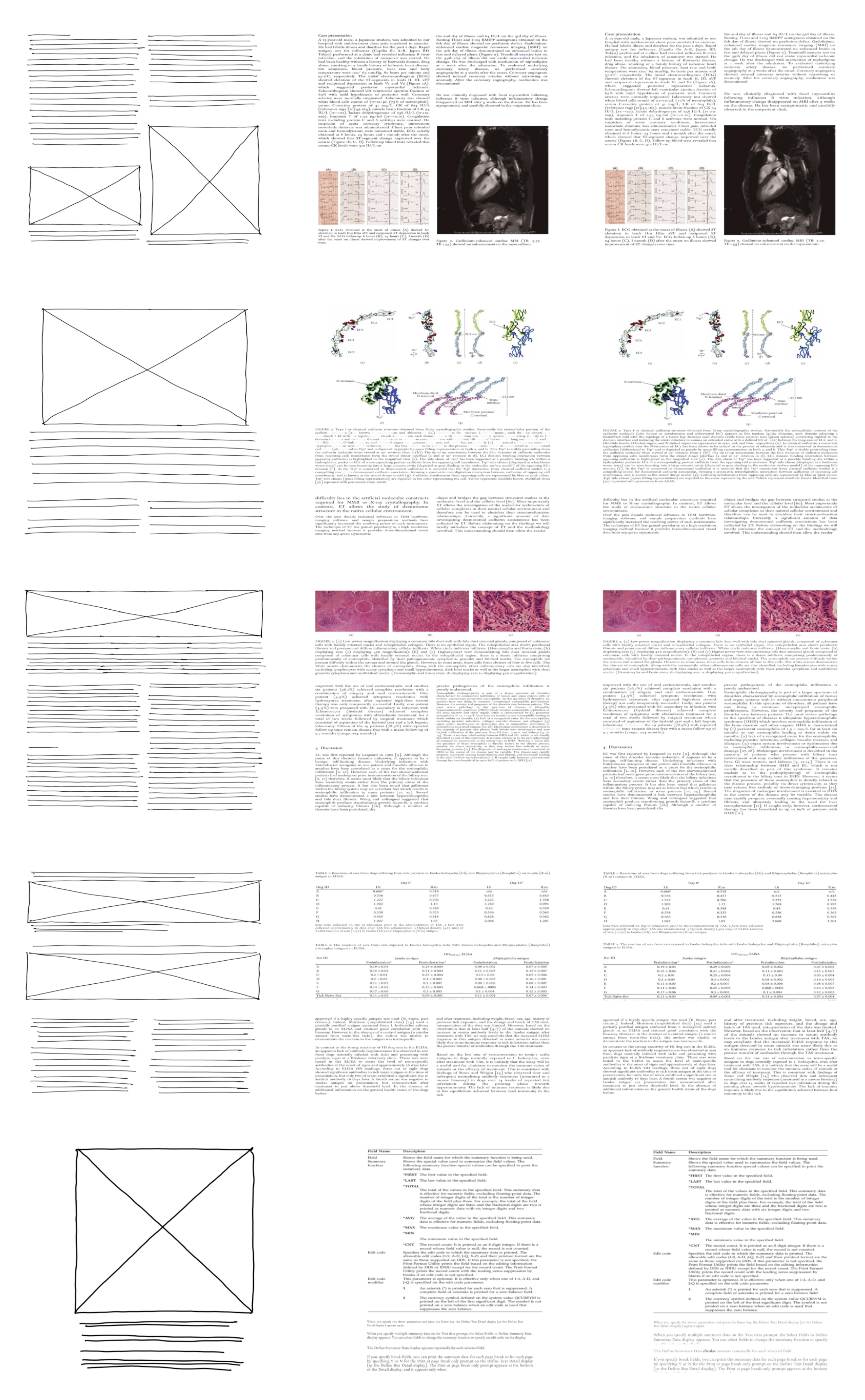}
      \put(8, 99){ Sketch}
        \put(27, 99){ Prediction}
        \put(49, 99){ Target}

        \put(-2, 83){\rotatebox{90}{ PubLayNet Example\footnotemark[7]}}
        \put(-2, 63){\rotatebox{90}{ PubLayNet Example\footnotemark[8]}}
        \put(-2, 43){\rotatebox{90}{ PubLayNet Example\footnotemark[9]}}
        \put(-2, 23){\rotatebox{90}{ PubLayNet Example\footnotemark[6]}}
        \put(-2, 4){\rotatebox{90}{ DoclayNet Example\footnotemark[4]}}
    \end{overpic}
    \caption{Sketches with corresponding predictions and the target layouts. Our method is able to generate layouts that conform to the sketch and have meaningful semantic order.}
    \label{fig:rendered_examples1}
\end{figure*}

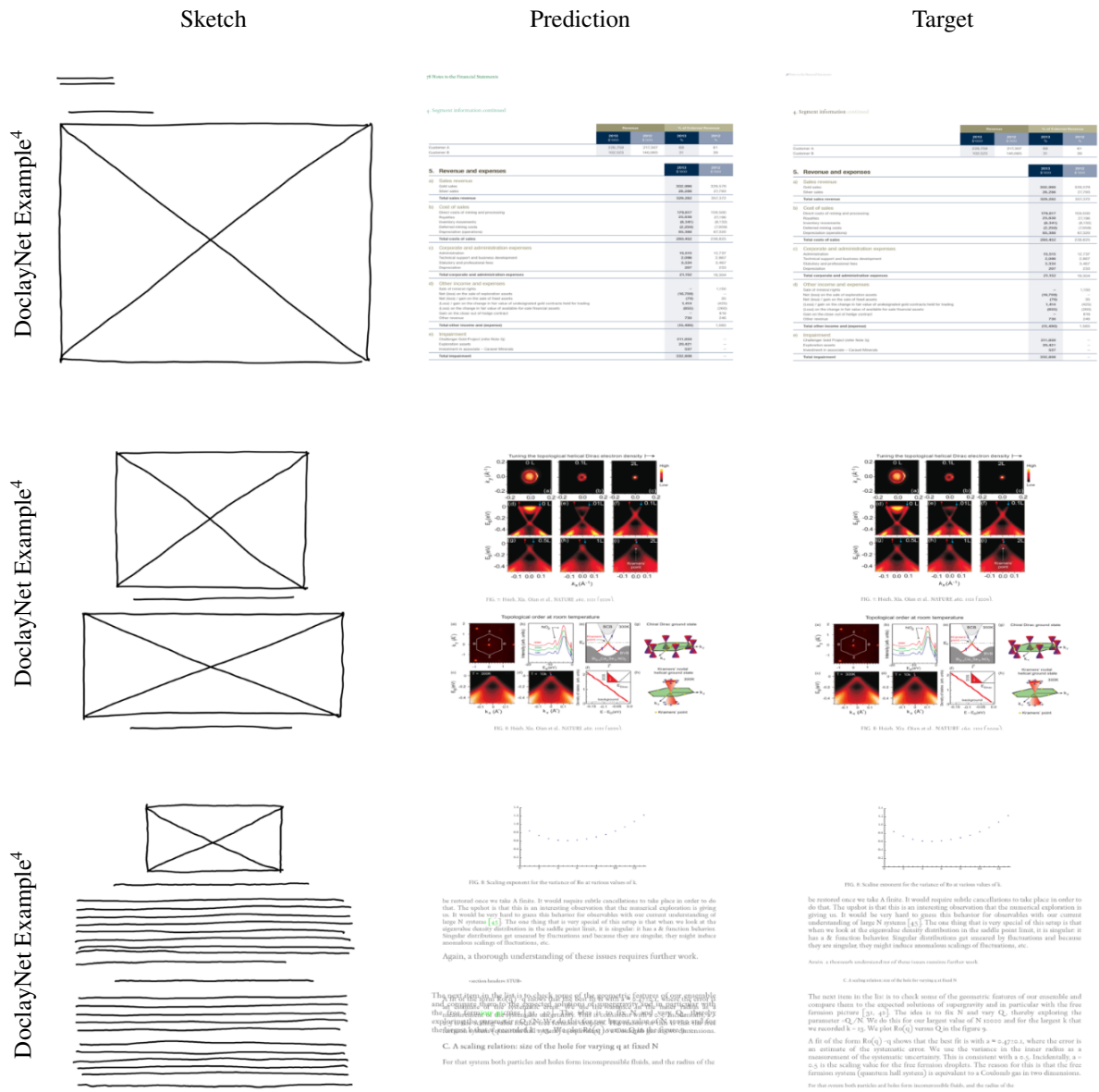
\begin{figure*}[htbp]
    \centering
    \begin{overpic}[width=0.9\textwidth,grid=false,tics=10]{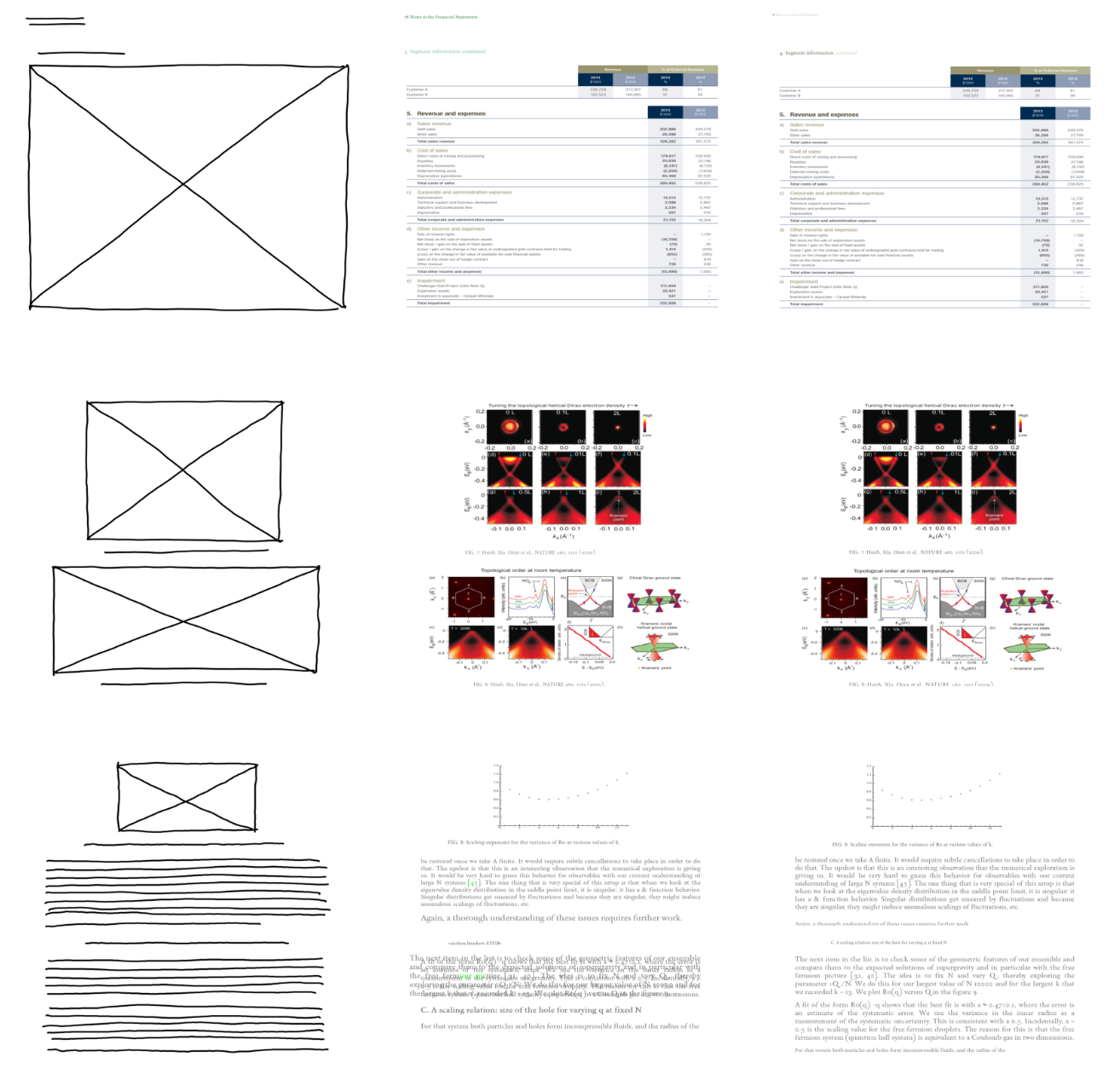}
     \put(13, 99){ Sketch}
        \put(45, 99){ Prediction}
        \put(80, 99){ Target}

        \put(-2, 70){\rotatebox{90}{ DoclayNet Example\footnotemark[4]}}
        \put(-2, 38){\rotatebox{90}{ DoclayNet Example\footnotemark[4]}}
        \put(-2, 4){\rotatebox{90}{ DoclayNet Example\footnotemark[4]}}
    \end{overpic}
    \caption{Sketches with corresponding predictions and the target layouts.}
    \label{fig:rendered_examples2}
\end{figure*}

\label{sec:rationale}

\clearpage
\newpage
\section{Legal Attributions}

\footnotetext[2]{Target image licensed under CC-BY license available at \href{http://creativecommons.org/licenses/by/2.0}{http://creativecommons.org/licenses/by/2.0}. Handschel J, Naujoks C, Depprich R, Lammers L, Kübler N, Meyer U, Wiesmann HP. Embryonic stem cells in scaffold-free three-dimensional cell culture: osteogenic differentiation and bone generation. Head Face Med. 2011 Jul 14;7:12. doi: 10.1186/1746-160X-7-12. PMID: 21752302; PMCID: PMC3143924}

\footnotetext[3]{Target image licensed under CC-BY license available at \href{http://creativecommons.org/licenses/by/4.0}{http://creativecommons.org/licenses/by/4.0}. Theruvath TP, Ramshesh VK, Zhong Z, Currin RT, Karrasch T, Lemasters JJ. Icam-1 upregulation in ethanol-induced Fatty murine livers promotes injury and sinusoidal leukocyte adherence after transplantation. HPB Surg. 2012;2012:480893. doi: 10.1155/2012/480893. Epub 2012 Jun 18. PMID: 22778492; PMCID: PMC3385666}

\footnotetext[4]{DocLayNet examples are licensed under CDLA - Permissive, Version 1.0, available at \href{https://cdla.dev/permissive-1-0/}{https://cdla.dev/permissive-1-0/}. DocLayNet: A Large Human-Annotated Dataset for Document-Layout Analysis Pfitzmann, Birgit and Auer, Christoph and Dolfi, Michele and Nassar, Ahmed S and Staar, Peter W J}

\footnotetext[5]{SlideVQA examples are licensed under CC-BY available at \href{https://creativecommons.org/licenses/by/4.0/}{https://creativecommons.org/licenses/by/4.0/}. SlideVQA: A Dataset for Document Visual Question Answering on Multiple Images, Tanaka, Ryota, Nishida, Kyosuke, Hasegawa, Taku, Saito, Itsumi, Saito, Kuniko, 2023/01/12}

\footnotetext[6]{Target image licensed under CC-BY license available at \href{http://creativecommons.org/licenses/by/4.0}{http://creativecommons.org/licenses/by/4.0}. Hall-Mendelin S, O'Donoghue P, Atwell RB, Lee R, Hall RA. An ELISA to Detect Serum Antibodies to the Salivary Gland Toxin of Ixodes holocyclus Neumann in Dogs and Rodents. J Parasitol Res. 2011;2011:283416. doi: 10.1155/2011/283416. Epub 2011 May 18. PMID: 21687655; PMCID: PMC3112514.}

\footnotetext[7]{Target image licensed under CC-BY license available at \href{http://creativecommons.org/licenses/by/3.0}{http://creativecommons.org/licenses/by/3.0}. Muneuchi J, Kanaya Y, Takimoto T, Hoshina T, Kusuhara K, Hara T. Myocarditis mimicking acute coronary syndrome following influenza B virus infection: a case report. Cases J. 2009 Jun 25;2:6809. doi: 10.4076/1757-1626-2-6809. PMID: 19829864; PMCID: PMC2740055.}

\footnotetext[8]{Target image licensed under CC-BY license available at \href{https://creativecommons.org/licenses/by/4.0/}{https://creativecommons.org/licenses/by/4.0/}. Owen GR, Stokes DL. Exploring the Nature of Desmosomal Cadherin Associations in 3D. Dermatol Res Pract. 2010;2010:930401. doi: 10.1155/2010/930401. Epub 2010 Jun 21. PMID: 20672011; PMCID: PMC2905946.}

\footnotetext[9]{Target image licensed under CC-BY license available at \href{http://creativecommons.org/licenses/by/4.0}{http://creativecommons.org/licenses/by/4.0}. Nashed C, Sakpal SV, Shusharina V, Chamberlain RS. Eosinophilic cholangitis and cholangiopathy: a sheep in wolves clothing. HPB Surg. 2010;2010:906496. doi: 10.1155/2010/906496. Epub 2010 Nov 7. PMID: 21076681; PMCID: PMC2976516.}

%% file: main.bbl
\begin{thebibliography}{45}
\providecommand{\natexlab}[1]{#1}
\providecommand{\url}[1]{\texttt{#1}}
\expandafter\ifx\csname urlstyle\endcsname\relax
  \providecommand{\doi}[1]{doi: #1}\else
  \providecommand{\doi}{doi: \begingroup \urlstyle{rm}\Url}\fi

\bibitem[pro()]{protobuf}
Protocol buffer documentation.
\newblock \url{https://protobuf.dev/}.

\bibitem[Baul{\'e} et~al.(2021)Baul{\'e}, Von~Wangenheim, Von~Wangenheim,
  Hauck, and J{\'u}nior]{baule2021automatic}
Daniel Baul{\'e}, Christiane~Gresse Von~Wangenheim, Aldo Von~Wangenheim,
  Jean~CR Hauck, and Edson C~Vargas J{\'u}nior.
\newblock {Automatic code generation from sketches of mobile applications in
  end-user development using Deep Learning}.
\newblock \emph{arXiv preprint arXiv:2103.05704}, 2021.

\bibitem[Beyer et~al.(2024)Beyer, Steiner, Pinto, Kolesnikov, Wang, Salz,
  Neumann, Alabdulmohsin, Tschannen, Bugliarello, Unterthiner, Keysers,
  Koppula, Liu, Grycner, Gritsenko, Houlsby, Kumar, Rong, Eisenschlos, Kabra,
  Bauer, Bošnjak, Chen, Minderer, Voigtlaender, Bica, Balazevic, Puigcerver,
  Papalampidi, Henaff, Xiong, Soricut, Harmsen, and
  Zhai]{beyer2024paligemmaversatile3bvlm}
Lucas Beyer, Andreas Steiner, André~Susano Pinto, Alexander Kolesnikov, Xiao
  Wang, Daniel Salz, Maxim Neumann, Ibrahim Alabdulmohsin, Michael Tschannen,
  Emanuele Bugliarello, Thomas Unterthiner, Daniel Keysers, Skanda Koppula,
  Fangyu Liu, Adam Grycner, Alexey Gritsenko, Neil Houlsby, Manoj Kumar, Keran
  Rong, Julian Eisenschlos, Rishabh Kabra, Matthias Bauer, Matko Bošnjak, Xi
  Chen, Matthias Minderer, Paul Voigtlaender, Ioana Bica, Ivana Balazevic, Joan
  Puigcerver, Pinelopi Papalampidi, Olivier Henaff, Xi Xiong, Radu Soricut,
  Jeremiah Harmsen, and Xiaohua Zhai.
\newblock {PaliGemma: A versatile 3B VLM for transfer}, 2024.

\bibitem[Buxton(2010)]{buxton2010sketching}
Bill Buxton.
\newblock \emph{{Sketching User Experiences: Getting the Design Right and the
  Right Design}}.
\newblock Morgan kaufmann, 2010.

\bibitem[Carion et~al.(2020)Carion, Massa, Synnaeve, Usunier, Kirillov, and
  Zagoruyko]{carion2020end}
Nicolas Carion, Francisco Massa, Gabriel Synnaeve, Nicolas Usunier, Alexander
  Kirillov, and Sergey Zagoruyko.
\newblock {End-to-end Object Detection with Transformers}.
\newblock In \emph{European conference on computer vision}, pages 213--229.
  Springer, 2020.

\bibitem[Cheng et~al.(2023)Cheng, Huang, Li, and
  Li]{cheng2023playparametricallyconditionedlayout}
Chin-Yi Cheng, Forrest Huang, Gang Li, and Yang Li.
\newblock {PLay: Parametrically Conditioned Layout Generation using Latent
  Diffusion}.
\newblock In \emph{ICML}, 2023.

\bibitem[Cheng et~al.(2024)Cheng, Zhang, Yang, Nie, Li, Wu, and
  Shao]{cheng2024graphic}
Yutao Cheng, Zhao Zhang, Maoke Yang, Hui Nie, Chunyuan Li, Xinglong Wu, and Jie
  Shao.
\newblock {Graphic Design with Large Multimodal Model}.
\newblock \emph{arXiv preprint arXiv:2404.14368}, 2024.

\bibitem[Devlin et~al.(2019)Devlin, Chang, Lee, and
  Toutanova]{devlin2019bertpretrainingdeepbidirectional}
Jacob Devlin, Ming-Wei Chang, Kenton Lee, and Kristina Toutanova.
\newblock {BERT: Pre-training of Deep Bidirectional Transformers for Language
  Understanding}, 2019.

\bibitem[Dosovitskiy et~al.(2020)Dosovitskiy, Beyer, Kolesnikov, Weissenborn,
  Zhai, Unterthiner, Dehghani, Minderer, Heigold, Gelly,
  et~al.]{dosovitskiy2020image}
Alexey Dosovitskiy, Lucas Beyer, Alexander Kolesnikov, Dirk Weissenborn,
  Xiaohua Zhai, Thomas Unterthiner, Mostafa Dehghani, Matthias Minderer, Georg
  Heigold, Sylvain Gelly, et~al.
\newblock {An image is worth 16x16 words: Transformers for image recognition at
  scale}.
\newblock \emph{arXiv preprint arXiv:2010.11929}, 2020.

\bibitem[Ferreira et~al.(2021)Ferreira, Restivo, and
  Ferreira]{ferreira2021automatically}
Joao~Silva Ferreira, Andr{\'e} Restivo, and Hugo~Sereno Ferreira.
\newblock {Automatically Generating Websites from Hand-drawn Mockups}.
\newblock In \emph{Proceedings of the 16th International Joint Conference on
  Computer Vision, Imaging and Computer Graphics Theory and Applications},
  2021.

\bibitem[Gupta et~al.(2021)Gupta, Lazarow, Achille, Davis, Mahadevan, and
  Shrivastava]{gupta2021layouttransformer}
Kamal Gupta, Justin Lazarow, Alessandro Achille, Larry~S Davis, Vijay
  Mahadevan, and Abhinav Shrivastava.
\newblock {LayoutTransformer: Layout Generation and Completion with
  self-attention}.
\newblock In \emph{Proceedings of the IEEE/CVF International Conference on
  Computer Vision}, pages 1004--1014, 2021.

\bibitem[Hsu et~al.(2023)Hsu, He, Peng, Kong, and Zhang]{hsu2023posterlayout}
Hsiao~Yuan Hsu, Xiangteng He, Yuxin Peng, Hao Kong, and Qing Zhang.
\newblock {PosterLayout: A new Benchmark and Approach for Content-aware
  Visual-textual Presentation Layout}.
\newblock In \emph{Proceedings of the IEEE/CVF Conference on Computer Vision
  and Pattern Recognition}, pages 6018--6026, 2023.

\bibitem[Inoue et~al.(2023{\natexlab{a}})Inoue, Kikuchi, Simo-Serra, Otani, and
  Yamaguchi]{Inoue_2023_CVPR}
Naoto Inoue, Kotaro Kikuchi, Edgar Simo-Serra, Mayu Otani, and Kota Yamaguchi.
\newblock {LayoutDM: Discrete Diffusion Model for Controllable Layout
  Generation}.
\newblock In \emph{Proceedings of the IEEE/CVF Conference on Computer Vision
  and Pattern Recognition (CVPR)}, pages 10167--10176, 2023{\natexlab{a}}.

\bibitem[Inoue et~al.(2023{\natexlab{b}})Inoue, Kikuchi, Simo-Serra, Otani, and
  Yamaguchi]{inoue2023towards}
Naoto Inoue, Kotaro Kikuchi, Edgar Simo-Serra, Mayu Otani, and Kota Yamaguchi.
\newblock {Towards Flexible Multi-modal Document Models}.
\newblock In \emph{Proceedings of the IEEE/CVF Conference on Computer Vision
  and Pattern Recognition}, pages 14287--14296, 2023{\natexlab{b}}.

\bibitem[Jain et~al.(2019)Jain, Agrawal, Banga, Kapoor, and
  Gulyani]{jain2019sketch2code}
Vanita Jain, Piyush Agrawal, Subham Banga, Rishabh Kapoor, and Shashwat
  Gulyani.
\newblock {Sketch2Code: Transformation of Sketches to UI in Real-time using
  Deep Neural Network}.
\newblock \emph{arXiv preprint arXiv:1910.08930}, 2019.

\bibitem[Jiang et~al.(2023)Jiang, Guo, Sun, Deng, Wu, Mijovic, Yang, Lou, and
  Zhang]{jiang2023layoutformer++}
Zhaoyun Jiang, Jiaqi Guo, Shizhao Sun, Huayu Deng, Zhongkai Wu, Vuksan Mijovic,
  Zijiang~James Yang, Jian-Guang Lou, and Dongmei Zhang.
\newblock {LayoutFormer++: Conditional Graphic Layout Generation via Constraint
  Serialization and Decoding Space Restriction}.
\newblock In \emph{Proceedings of the IEEE/CVF Conference on Computer Vision
  and Pattern Recognition}, pages 18403--18412, 2023.

\bibitem[Jyothi et~al.(2019)Jyothi, Durand, He, Sigal, and
  Mori]{jyothi2019layoutvae}
Akash~Abdu Jyothi, Thibaut Durand, Jiawei He, Leonid Sigal, and Greg Mori.
\newblock {LayoutVAE: Stochastic Scene Layout Generation from a Label Set}.
\newblock In \emph{Proceedings of the IEEE/CVF International Conference on
  Computer Vision}, pages 9895--9904, 2019.

\bibitem[Kikuchi et~al.(2021)Kikuchi, Simo-Serra, Otani, and
  Yamaguchi]{Kikuchi_2021}
Kotaro Kikuchi, Edgar Simo-Serra, Mayu Otani, and Kota Yamaguchi.
\newblock {Constrained Graphic Layout Generation via Latent Optimization}.
\newblock In \emph{Proceedings of the 29th ACM International Conference on
  Multimedia}. ACM, 2021.

\bibitem[Lee et~al.(2020)Lee, Jiang, Essa, Le, Gong, Yang, and
  Yang]{lee2020neural}
Hsin-Ying Lee, Lu Jiang, Irfan Essa, Phuong~B Le, Haifeng Gong, Ming-Hsuan
  Yang, and Weilong Yang.
\newblock {Neural design network: Graphic layout generation with constraints}.
\newblock In \emph{Computer Vision--ECCV 2020: 16th European Conference,
  Glasgow, UK, August 23--28, 2020, Proceedings, Part III 16}, pages 491--506.
  Springer, 2020.

\bibitem[{Levenshtein}(1966)]{levenshteindistance}
V.~I. {Levenshtein}.
\newblock {Binary Codes Capable of Correcting Deletions, Insertions and
  Reversals}.
\newblock \emph{Soviet Physics Doklady}, 10:\penalty0 707, 1966.

\bibitem[Li et~al.(2020{\natexlab{a}})Li, Yang, Hertzmann, Zhang, and
  Xu]{li2020layoutgan}
Jianan Li, Jimei Yang, Aaron Hertzmann, Jianming Zhang, and Tingfa Xu.
\newblock {LayoutGAN: Synthesizing Graphic Layouts with Vector-wireframe
  Adversarial Networks}.
\newblock \emph{IEEE Transactions on Pattern Analysis and Machine
  Intelligence}, 43\penalty0 (7):\penalty0 2388--2399, 2020{\natexlab{a}}.

\bibitem[Li et~al.(2020{\natexlab{b}})Li, Yang, Zhang, Liu, Wang, and
  Xu]{li2020attributeconditionedlayoutganautomatic}
Jianan Li, Jimei Yang, Jianming Zhang, Chang Liu, Christina Wang, and Tingfa
  Xu.
\newblock {Attribute-conditioned Layout GAN for Automatic Graphic Design},
  2020{\natexlab{b}}.

\bibitem[Li et~al.(2024)Li, Zhang, and Yang]{li2024sketch2code}
Ryan Li, Yanzhe Zhang, and Diyi Yang.
\newblock {Sketch2Code: Evaluating Vision-Language Models for Interactive Web
  Design Prototyping}.
\newblock \emph{arXiv preprint arXiv:2410.16232}, 2024.

\bibitem[Liang and Lin(2023)]{liang2023sketch2wireframe}
Xudong Liang and Tao Lin.
\newblock {Sketch2Wireframe: an automatic framework for transforming hand-drawn
  sketches to digital wireframes in UI design}.
\newblock \emph{The Visual Computer}, pages 1--11, 2023.

\bibitem[Lin et~al.(2023)Lin, Guo, Sun, Xu, Liu, Lou, and Zhang]{lin2023parse}
Jiawei Lin, Jiaqi Guo, Shizhao Sun, Weijiang Xu, Ting Liu, Jian-Guang Lou, and
  Dongmei Zhang.
\newblock {A Parse-Then-Place Approach for Generating Graphic Layouts from
  Textual Descriptions}.
\newblock In \emph{Proceedings of the IEEE/CVF International Conference on
  Computer Vision}, pages 23622--23631, 2023.

\bibitem[Lin et~al.(2024)Lin, Guo, Sun, Yang, Lou, and
  Zhang]{lin2024layoutprompter}
Jiawei Lin, Jiaqi Guo, Shizhao Sun, Zijiang Yang, Jian-Guang Lou, and Dongmei
  Zhang.
\newblock {LayoutPrompter: Awaken the Design Ability of Large Language Models}.
\newblock \emph{Advances in Neural Information Processing Systems}, 36, 2024.

\bibitem[Loshchilov and Hutter(2017)]{cosinelrscheduler}
Ilya Loshchilov and Frank Hutter.
\newblock {SGDR: Stochastic Gradient Descent with Warm Restarts}, 2017.

\bibitem[Maneewongvatana and Mount(1999)]{maneewongvatana1999analysis}
Songrit Maneewongvatana and David~M Mount.
\newblock {Analysis of Approximate Nearest Neighbor Searching with Clustered
  Point Sets}.
\newblock \emph{arXiv preprint cs/9901013}, 1999.

\bibitem[Mohian and Csallner(2020)]{mohian2020doodle2app}
Soumik Mohian and Christoph Csallner.
\newblock {Doodle2App: Native app code by freehand UI sketching}.
\newblock In \emph{Proceedings of the IEEE/ACM 7th International Conference on
  Mobile Software Engineering and Systems}, pages 81--84, 2020.

\bibitem[Myers et~al.(2008)Myers, Park, Nakano, Mueller, and
  Ko]{myers2008designers}
Brad Myers, Sun~Young Park, Yoko Nakano, Greg Mueller, and Amy Ko.
\newblock {How Designers Design and Program Interactive Behaviors}.
\newblock In \emph{2008 IEEE Symposium on Visual Languages and Human-Centric
  Computing}, pages 177--184. IEEE, 2008.

\bibitem[Newman and Landay(2000)]{newman2000sitemaps}
Mark~W Newman and James~A Landay.
\newblock {Sitemaps, Storyboards, and Specifications: A Sketch of Web Site
  Design Practice}.
\newblock In \emph{Proceedings of the 3rd conference on Designing interactive
  systems: processes, practices, methods, and techniques}, pages 263--274,
  2000.

\bibitem[Pfitzmann et~al.(2022)Pfitzmann, Auer, Dolfi, Nassar, and
  Staar]{PfitzmannDoclaynet2022}
Birgit Pfitzmann, Christoph Auer, Michele Dolfi, Ahmed~S. Nassar, and Peter
  Staar.
\newblock {DocLayNet: A Large Human-Annotated Dataset for Document-Layout
  Segmentation}.
\newblock In \emph{Proceedings of the 28th ACM SIGKDD Conference on Knowledge
  Discovery and Data Mining}. ACM, 2022.

\bibitem[Seol et~al.(2024)Seol, Kim, and Yoo]{seol2024posterllama}
Jaejung Seol, Seojun Kim, and Jaejun Yoo.
\newblock {PosterLlama: Bridging Design Ability of Language Model to
  Contents-Aware Layout Generation}.
\newblock \emph{arXiv preprint arXiv:2404.00995}, 2024.

\bibitem[Shabani et~al.(2024)Shabani, Wang, Liu, Zhao, Yang, and
  Furukawa]{shabani2024visual}
Mohammad~Amin Shabani, Zhaowen Wang, Difan Liu, Nanxuan Zhao, Jimei Yang, and
  Yasutaka Furukawa.
\newblock {Visual Layout Composer: Image-Vector Dual Diffusion Model for Design
  Layout Generation}.
\newblock In \emph{Proceedings of the IEEE/CVF Conference on Computer Vision
  and Pattern Recognition}, pages 9222--9231, 2024.

\bibitem[Tanaka et~al.(2023)Tanaka, Nishida, Nishida, Hasegawa, Saito, and
  Saito]{tanaka2023slidevqadatasetdocumentvisual}
Ryota Tanaka, Kyosuke Nishida, Kosuke Nishida, Taku Hasegawa, Itsumi Saito, and
  Kuniko Saito.
\newblock {SlideVQA: A Dataset for Document Visual Question Answering on
  Multiple Images}, 2023.

\bibitem[Tang et~al.(2023)Tang, Wu, Li, and Duan]{tang2023layoutnuwa}
Zecheng Tang, Chenfei Wu, Juntao Li, and Nan Duan.
\newblock {LayoutNUWA: Revealing the Hidden Layout Expertise of Large Language
  Models}.
\newblock \emph{arXiv preprint arXiv:2309.09506}, 2023.

\bibitem[Team et~al.(2024{\natexlab{a}})Team, Georgiev, Lei, Burnell, Bai,
  Gulati, Tanzer, Vincent, Pan, Wang, Mariooryad, Ding, Geng, Alcober, Frostig,
  Omernick, Walker, Paduraru, Sorokin, Tacchetti, Gaffney, Daruki, Sercinoglu,
  Gleicher, Love, Voigtlaender, Jain, Surita, Mohamed, Blevins, Ahn, Zhu,
  Kawintiranon, Firat, Gu, Zhang, Rahtz, Faruqui, Clay, Gilmer, Co-Reyes,
  Penchev, Zhu, Morioka, Hui, Haridasan, Campos, Mahdieh, Guo, Hassan, Kilgour,
  Vezer, Cheng, de~Liedekerke, Goyal, Barham, Strouse, Noury, Adler,
  Sundararajan, Vikram, Lepikhin, Paganini, Garcia, Yang, Valter, Trebacz,
  Vodrahalli, Asawaroengchai, Ring, Kalb, Soares, Brahma, Steiner, Yu, Mentzer,
  He, Gonzalez, Xu, Kaufman, Shafey, Oh, Hennigan, van~den Driessche, Odoom,
  Lucic, Roelofs, Lall, Marathe, Chan, Ontanon, He, Teplyashin, Lai, Crone,
  Damoc, Ho, Riedel, Lenc, Yeh, Chowdhery, Xu, Kazemi, Amid, Petrushkina,
  Swersky, Khodaei, Chen, Larkin, Pinto, Yan, Badia, Patil, Hansen, Orr,
  Arnold, Grimstad, Dai, Douglas, Sinha, Yadav, Chen, Gribovskaya, Austin,
  Zhao, Patel, Komarek, Austin, Borgeaud, Friso, Goyal, Caine, Cao, Chung,
  Lamm, Barth-Maron, Kagohara, Olszewska, Chen, Shivakumar, Agarwal, Godhia,
  Rajwar, Snaider, Dotiwalla, Liu, Barua, Ungureanu, Zhang, Batsaikhan, Wirth,
  Qin, Danihelka, Doshi, Chadwick, Chen, Jain, Le, Kar, Gurumurthy, Li, Sang,
  Liu, Lamprou, Munoz, Lintz, Mehta, Howard, Reynolds, Aroyo, Wang, Blanco,
  Cassirer, Griffith, Das, Lee, Sygnowski, Fisher, Besley, Powell, Ahmed,
  Paulus, Reitter, Borsos, Joshi, Pope, Hand, Selo, Jain, Sethi, Goel, Makino,
  May, Yang, Schalkwyk, Butterfield, Hauth, Goldin, Hawkins, Senter, Brin,
  Woodman, Ritter, Noland, Giang, Bolina, Lee, Blyth, Mackinnon, Reid, Sarvana,
  Silver, Chen, Wang, Maggiore, Chang, Attaluri, Thornton, Chiu, Bunyan,
  Levine, Chung, Eltyshev, Si, Lillicrap, Brady, Aggarwal, Wu, Xu, McIlroy,
  Badola, Sandhu, Moreira, Stokowiec, Hemsley, Li, Tudor, Shyam, Rahimtoroghi,
  Haykal, Sprechmann, Zhou, Mincu, Li, Addanki, Krishna, Wu, Frechette, Eyal,
  Dafoe, Lacey, Whang, Avrahami, Zhang, Taropa, Lin, Toyama, Rutherford, Sano,
  Choe, Tomala, Safranek-Shrader, Kassner, Pajarskas, Harvey, Sechrist,
  Fortunato, Lyu, Elsayed, Kuang, Lottes, Chu, Jia, Chen, Humphreys, Baumli,
  Tao, Samuel, dos Santos, Andreassen, Rakićević, Grewe, Kumar, Winkler,
  Caton, Brock, Dalmia, Sheahan, Barr, Miao, Natsev, Devlin, Behbahani, Prost,
  Sun, Myaskovsky, Pillai, Hurt, Lazaridou, Xiong, Zheng, Pardo, Li, Horgan,
  Stanton, Ambar, Xia, Lince, Wang, Mustafa, Webson, Lee, Anil, Wicke, Dozat,
  Sinha, Piqueras, Dabir, Upadhyay, Boral, Hendricks, Fry, Djolonga, Su,
  Walker, Labanowski, Huang, Misra, Chen, Skerry-Ryan, Singh, Rijhwani, Yu,
  Castro-Ros, Changpinyo, Datta, Bagri, Hrafnkelsson, Maggioni, Zheng, Sulsky,
  Hou, Paine, Yang, Riesa, Rogozinska, Marcus, Badawy, Zhang, Wang, Miller,
  Greer, Sjos, Nova, Zen, Chaabouni, Rosca, Jiang, Chen, Liu, Sainath, Krikun,
  Polozov, Lespiau, Newlan, Cankara, Kwak, Xu, Chen, Coenen, Meyer, Tsihlas,
  Ma, Gottweis, Xing, Gu, Miao, Frank, Cankara, Ganapathy, Dasgupta,
  Hughes-Fitt, Chen, Reid, Rong, Fan, van Amersfoort, Zhuang, Cohen, Gu,
  Mohananey, Ilic, Tobin, Wieting, Bortsova, Thacker, Wang, Caveness, Chiu,
  Sezener, Kaskasoli, Baker, Millican, Elhawaty, Aisopos, Lebsack, Byrd, Dai,
  Jia, Wiethoff, Davoodi, Weston, Yagati, Ahuja, Gao, Pundak, Zhang, Azzam,
  Sim, Caelles, Keeling, Sharma, Swing, Li, Liu, Bostock, Bansal, Nado, Anand,
  Lipschultz, Karmarkar, Proleev, Ittycheriah, Yeganeh, Polovets, Faust, Sun,
  Rrustemi, Li, Shivanna, Liu, Welty, Lebron, Baddepudi, Krause, Parisotto,
  Soricut, Xu, Bloxwich, Johnson, Neyshabur, Mao-Jones, Wang, Ramasesh, Abbas,
  Guez, Segal, Nguyen, Svensson, Hou, York, Milan, Bridgers, Gworek,
  Tagliasacchi, Lee-Thorp, Chang, Guseynov, Hartman, Kwong, Zhao, Kashem, Cole,
  Miech, Tanburn, Phuong, Pavetic, Cevey, Comanescu, Ives, Yang, Du, Li, Zhang,
  Iinuma, Hu, Roy, Bijwadia, Zhu, Martins, Saputro, Gergely, Zheng, Jia,
  Antonoglou, Sadovsky, Gu, Bi, Andreev, Samangooei, Khan, Kocisky, Filos,
  Kumar, Bishop, Yu, Hodkinson, Mittal, Shah, Moufarek, Cheng, Bloniarz, Lee,
  Pejman, Michel, Spencer, Feinberg, Xiong, Savinov, Smith, Shakeri, Tran,
  Chesus, Bohnet, Tucker, von Glehn, Muir, Mao, Kazawa, Slone, Soparkar,
  Shrivastava, Cobon-Kerr, Sharman, Pavagadhi, Araya, Misiunas, Ghelani,
  Laskin, Barker, Li, Briukhov, Houlsby, Glaese, Lakshminarayanan, Schucher,
  Tang, Collins, Lim, Feng, Recasens, Lai, Magni, Cao, Siddhant, Ashwood,
  Orbay, Dehghani, Brennan, He, Xu, Gao, Saroufim, Molloy, Wu, Arnold, Chang,
  Schrittwieser, Buchatskaya, Radpour, Polacek, Giordano, Bapna, Tokumine,
  Hellendoorn, Sottiaux, Cogan, Severyn, Saleh, Thakoor, Shefey, Qiao, Gaba,
  yiin Chang, Swanson, Zhang, Lee, Rubenstein, Song, Kwiatkowski, Koop, Kannan,
  Kao, Schuh, Stjerngren, Ghiasi, Gibson, Vilnis, Yuan, Ferreira, Kamath,
  Klimenko, Franko, Xiao, Bhattacharya, Patel, Wang, Morris, Strudel, Sharma,
  Choy, Hashemi, Landon, Finkelstein, Jhakra, Frye, Barnes, Mauger, Daun,
  Baatarsukh, Tung, Farhan, Michalewski, Viola, de~Chaumont~Quitry, Lan,
  Hudson, Wang, Fischer, Zheng, White, Dragan, baptiste Alayrac, Ni, Pritzel,
  Iwanicki, Isard, Bulanova, Zilka, Dyer, Sachan, Srinivasan, Muckenhirn, Cai,
  Mandhane, Tariq, Rae, Wang, Ayoub, FitzGerald, Zhao, Han, Alberti, Garrette,
  Krishnakumar, Gimenez, Levskaya, Sohn, Matak, Iturrate, Chang, Xiang, Cao,
  Ranka, Brown, Hutter, Mirrokni, Chen, Yao, Egyed, Galilee, Liechty,
  Kallakuri, Palmer, Ghemawat, Liu, Tao, Thornton, Green, Jasarevic, Lin,
  Cotruta, Tan, Fiedel, Yu, Chi, Neitz, Heitkaemper, Sinha, Zhou, Sun, Kaed,
  Hulse, Mishra, Georgaki, Kudugunta, Farabet, Shafran, Vlasic, Tsitsulin,
  Ananthanarayanan, Carin, Su, Sun, V, Carvajal, Broder, Comsa, Repina, Wong,
  Chen, Hawkins, Filonov, Loher, Hirnschall, Wang, Ye, Burns, Cate, Wright,
  Piccinini, Zhang, Lin, Gog, Kulizhskaya, Sreevatsa, Song, Cobo, Iyer, Tekur,
  Garrido, Xiao, Kemp, Zheng, Li, Agarwal, Ngani, Goshvadi,
  Santamaria-Fernandez, Fica, Chen, Gorgolewski, Sun, Garg, Ye, Eslami, Hua,
  Simon, Joshi, Kim, Tenney, Potluri, Thiet, Yuan, Luisier, Chronopoulou,
  Scellato, Srinivasan, Chen, Koverkathu, Dalibard, Xu, Saeta, Anderson,
  Sellam, Fernando, Huot, Jung, Varadarajan, Quinn, Raul, Le, Habalov, Clark,
  Jalan, Bullard, Singhal, Luong, Wang, Rajayogam, Eisenschlos, Jia,
  Finchelstein, Yakubovich, Balle, Fink, Agarwal, Li, Dvijotham, Pal, Kang,
  Konzelmann, Beattie, Dousse, Wu, Crocker, Elkind, Jonnalagadda, Lee,
  Holtmann-Rice, Kallarackal, Liu, Vnukov, Vats, Invernizzi, Jafari, Zhou,
  Taylor, Prendki, Wu, Eccles, Liu, Kopparapu, Beaufays, Angermueller, Marzoca,
  Sarcar, Dib, Stanway, Perbet, Trdin, Sterneck, Khorlin, Li, Wu, Goenka,
  Madras, Goldshtein, Gierke, Zhou, Liu, Liang, White, Li, Singh, Bahargam,
  Epstein, Basu, Lao, Ozturel, Crous, Zhai, Lu, Tung, Gaur, Walton, Dixon,
  Zhang, Globerson, Uy, Bolt, Wiles, Nasr, Shumailov, Selvi, Piccinno, Aguilar,
  McCarthy, Khalman, Shukla, Galic, Carpenter, Villela, Zhang, Richardson,
  Martens, Bosnjak, Belle, Seibert, Alnahlawi, McWilliams, Singh, Louis, Ding,
  Popovici, Simicich, Knight, Mehta, Gupta, Shi, Fatehi, Mitrovic, Grills,
  Pagadora, Petrova, Eisenbud, Zhang, Yates, Mittal, Tripuraneni, Assael,
  Brovelli, Jain, Velimirovic, Akbulut, Mu, Macherey, Kumar, Xu, Qureshi,
  Comanici, Wiesner, Gong, Ruddock, Bauer, Felt, GP, Arnab, Zelle, Rothfuss,
  Rosgen, Shenoy, Seybold, Li, Mudigonda, Erdogan, Xia, Simsa, Michi, Yao, Yew,
  Kan, Caswell, Radebaugh, Elisseeff, Valenzuela, McKinney, Paterson, Cui,
  Latorre-Chimoto, Kim, Zeng, Durden, Ponnapalli, Sosea, Choquette-Choo,
  Manyika, Robenek, Vashisht, Pereira, Lam, Velic, Owusu-Afriyie, Lee,
  Bolukbasi, Parrish, Lu, Park, Venkatraman, Talbert, Rosique, Cheng,
  Sozanschi, Paszke, Kumar, Austin, Li, Salama, Kim, Dukkipati, Baryshnikov,
  Kaplanis, Sheng, Chervonyi, Unlu, de~Las~Casas, Askham, Tunyasuvunakool,
  Gimeno, Poder, Kwak, Miecnikowski, Mirrokni, Dimitriev, Parisi, Liu, Tsai,
  Shevlane, Kouridi, Garmon, Goedeckemeyer, Brown, Vijayakumar, Elqursh,
  Jazayeri, Huang, Carthy, Hoover, Kim, Kumar, Chen, Biles, Bingham, Rosen,
  Wang, Tan, Engel, Pongetti, de~Cesare, Hwang, Yu, Pullman, Narayanan, Levin,
  Gopal, Li, Aharoni, Trinh, Lo, Casagrande, Vij, Matthey, Ramadhana, Matthews,
  Carey, Johnson, Goranova, Shah, Ashraf, Dasgupta, Larsen, Wang, Vuyyuru,
  Jiang, Ijazi, Osawa, Smith, Boppana, Bilal, Koizumi, Xu, Altun, Shabat,
  Bariach, Korchemniy, Choo, Ronneberger, Iwuanyanwu, Zhao, Soergel, Hsieh,
  Cai, Iqbal, Sundermeyer, Chen, Bursztein, Malaviya, Biadsy, Shroff, Dhillon,
  Latkar, Dyer, Forbes, Nicosia, Nikolaev, Greene, Georgiev, Wang, Martin,
  Sedghi, Zhang, Banzal, Fritz, Rao, Wang, Zhang, Patraucean, Du, Mordatch,
  Jurin, Liu, Dubey, Mohan, Nowakowski, Ion, Wei, Tojo, Raad, Hudson, Keshava,
  Agrawal, Ramirez, Wu, Nguyen, Liu, Sewak, Petrini, Choi, Philips, Wang, Bica,
  Garg, Wilkiewicz, Agrawal, Li, Guo, Xue, Shaik, Leach, Khan, Wiesinger,
  Jerome, Chakladar, Wang, Ornduff, Abu, Ghaffarkhah, Wainwright, Cortes, Liu,
  Maynez, Terzis, Samangouei, Mansour, Kępa, Aubet, Algymr, Banica, Weisz,
  Orban, Senges, Andrejczuk, Geller, Santo, Anklin, Merey, Baeuml, Strohman,
  Bai, Petrov, Wu, Hassabis, Kavukcuoglu, Dean, and
  Vinyals]{gemini15multimodal}
Gemini Team, Petko Georgiev, Ving~Ian Lei, Ryan Burnell, Libin Bai, Anmol
  Gulati, Garrett Tanzer, Damien Vincent, Zhufeng Pan, Shibo Wang, Soroosh
  Mariooryad, Yifan Ding, Xinyang Geng, Fred Alcober, Roy Frostig, Mark
  Omernick, Lexi Walker, Cosmin Paduraru, Christina Sorokin, Andrea Tacchetti,
  Colin Gaffney, Samira Daruki, Olcan Sercinoglu, Zach Gleicher, Juliette Love,
  Paul Voigtlaender, Rohan Jain, Gabriela Surita, Kareem Mohamed, Rory Blevins,
  Junwhan Ahn, Tao Zhu, Kornraphop Kawintiranon, Orhan Firat, Yiming Gu, Yujing
  Zhang, Matthew Rahtz, Manaal Faruqui, Natalie Clay, Justin Gilmer, JD
  Co-Reyes, Ivo Penchev, Rui Zhu, Nobuyuki Morioka, Kevin Hui, Krishna
  Haridasan, Victor Campos, Mahdis Mahdieh, Mandy Guo, Samer Hassan, Kevin
  Kilgour, Arpi Vezer, Heng-Tze Cheng, Raoul de Liedekerke, Siddharth Goyal,
  Paul Barham, DJ Strouse, Seb Noury, Jonas Adler, Mukund Sundararajan, Sharad
  Vikram, Dmitry Lepikhin, Michela Paganini, Xavier Garcia, Fan Yang, Dasha
  Valter, Maja Trebacz, Kiran Vodrahalli, Chulayuth Asawaroengchai, Roman Ring,
  Norbert Kalb, Livio~Baldini Soares, Siddhartha Brahma, David Steiner, Tianhe
  Yu, Fabian Mentzer, Antoine He, Lucas Gonzalez, Bibo Xu, Raphael~Lopez
  Kaufman, Laurent~El Shafey, Junhyuk Oh, Tom Hennigan, George van~den
  Driessche, Seth Odoom, Mario Lucic, Becca Roelofs, Sid Lall, Amit Marathe,
  Betty Chan, Santiago Ontanon, Luheng He, Denis Teplyashin, Jonathan Lai, Phil
  Crone, Bogdan Damoc, Lewis Ho, Sebastian Riedel, Karel Lenc, Chih-Kuan Yeh,
  Aakanksha Chowdhery, Yang Xu, Mehran Kazemi, Ehsan Amid, Anastasia
  Petrushkina, Kevin Swersky, Ali Khodaei, Gowoon Chen, Chris Larkin, Mario
  Pinto, Geng Yan, Adria~Puigdomenech Badia, Piyush Patil, Steven Hansen, Dave
  Orr, Sebastien M.~R. Arnold, Jordan Grimstad, Andrew Dai, Sholto Douglas,
  Rishika Sinha, Vikas Yadav, Xi Chen, Elena Gribovskaya, Jacob Austin, Jeffrey
  Zhao, Kaushal Patel, Paul Komarek, Sophia Austin, Sebastian Borgeaud, Linda
  Friso, Abhimanyu Goyal, Ben Caine, Kris Cao, Da-Woon Chung, Matthew Lamm,
  Gabe Barth-Maron, Thais Kagohara, Kate Olszewska, Mia Chen, Kaushik
  Shivakumar, Rishabh Agarwal, Harshal Godhia, Ravi Rajwar, Javier Snaider,
  Xerxes Dotiwalla, Yuan Liu, Aditya Barua, Victor Ungureanu, Yuan Zhang,
  Bat-Orgil Batsaikhan, Mateo Wirth, James Qin, Ivo Danihelka, Tulsee Doshi,
  Martin Chadwick, Jilin Chen, Sanil Jain, Quoc Le, Arjun Kar, Madhu
  Gurumurthy, Cheng Li, Ruoxin Sang, Fangyu Liu, Lampros Lamprou, Rich Munoz,
  Nathan Lintz, Harsh Mehta, Heidi Howard, Malcolm Reynolds, Lora Aroyo, Quan
  Wang, Lorenzo Blanco, Albin Cassirer, Jordan Griffith, Dipanjan Das, Stephan
  Lee, Jakub Sygnowski, Zach Fisher, James Besley, Richard Powell, Zafarali
  Ahmed, Dominik Paulus, David Reitter, Zalan Borsos, Rishabh Joshi, Aedan
  Pope, Steven Hand, Vittorio Selo, Vihan Jain, Nikhil Sethi, Megha Goel,
  Takaki Makino, Rhys May, Zhen Yang, Johan Schalkwyk, Christina Butterfield,
  Anja Hauth, Alex Goldin, Will Hawkins, Evan Senter, Sergey Brin, Oliver
  Woodman, Marvin Ritter, Eric Noland, Minh Giang, Vijay Bolina, Lisa Lee, Tim
  Blyth, Ian Mackinnon, Machel Reid, Obaid Sarvana, David Silver, Alexander
  Chen, Lily Wang, Loren Maggiore, Oscar Chang, Nithya Attaluri, Gregory
  Thornton, Chung-Cheng Chiu, Oskar Bunyan, Nir Levine, Timothy Chung, Evgenii
  Eltyshev, Xiance Si, Timothy Lillicrap, Demetra Brady, Vaibhav Aggarwal, Boxi
  Wu, Yuanzhong Xu, Ross McIlroy, Kartikeya Badola, Paramjit Sandhu, Erica
  Moreira, Wojciech Stokowiec, Ross Hemsley, Dong Li, Alex Tudor, Pranav Shyam,
  Elahe Rahimtoroghi, Salem Haykal, Pablo Sprechmann, Xiang Zhou, Diana Mincu,
  Yujia Li, Ravi Addanki, Kalpesh Krishna, Xiao Wu, Alexandre Frechette, Matan
  Eyal, Allan Dafoe, Dave Lacey, Jay Whang, Thi Avrahami, Ye Zhang, Emanuel
  Taropa, Hanzhao Lin, Daniel Toyama, Eliza Rutherford, Motoki Sano, HyunJeong
  Choe, Alex Tomala, Chalence Safranek-Shrader, Nora Kassner, Mantas Pajarskas,
  Matt Harvey, Sean Sechrist, Meire Fortunato, Christina Lyu, Gamaleldin
  Elsayed, Chenkai Kuang, James Lottes, Eric Chu, Chao Jia, Chih-Wei Chen,
  Peter Humphreys, Kate Baumli, Connie Tao, Rajkumar Samuel, Cicero~Nogueira
  dos Santos, Anders Andreassen, Nemanja Rakićević, Dominik Grewe, Aviral
  Kumar, Stephanie Winkler, Jonathan Caton, Andrew Brock, Sid Dalmia, Hannah
  Sheahan, Iain Barr, Yingjie Miao, Paul Natsev, Jacob Devlin, Feryal
  Behbahani, Flavien Prost, Yanhua Sun, Artiom Myaskovsky,
  Thanumalayan~Sankaranarayana Pillai, Dan Hurt, Angeliki Lazaridou, Xi Xiong,
  Ce Zheng, Fabio Pardo, Xiaowei Li, Dan Horgan, Joe Stanton, Moran Ambar, Fei
  Xia, Alejandro Lince, Mingqiu Wang, Basil Mustafa, Albert Webson, Hyo Lee,
  Rohan Anil, Martin Wicke, Timothy Dozat, Abhishek Sinha, Enrique Piqueras,
  Elahe Dabir, Shyam Upadhyay, Anudhyan Boral, Lisa~Anne Hendricks, Corey Fry,
  Josip Djolonga, Yi Su, Jake Walker, Jane Labanowski, Ronny Huang, Vedant
  Misra, Jeremy Chen, RJ Skerry-Ryan, Avi Singh, Shruti Rijhwani, Dian Yu, Alex
  Castro-Ros, Beer Changpinyo, Romina Datta, Sumit Bagri, Arnar~Mar
  Hrafnkelsson, Marcello Maggioni, Daniel Zheng, Yury Sulsky, Shaobo Hou,
  Tom~Le Paine, Antoine Yang, Jason Riesa, Dominika Rogozinska, Dror Marcus,
  Dalia~El Badawy, Qiao Zhang, Luyu Wang, Helen Miller, Jeremy Greer, Lars~Lowe
  Sjos, Azade Nova, Heiga Zen, Rahma Chaabouni, Mihaela Rosca, Jiepu Jiang,
  Charlie Chen, Ruibo Liu, Tara Sainath, Maxim Krikun, Alex Polozov,
  Jean-Baptiste Lespiau, Josh Newlan, Zeyncep Cankara, Soo Kwak, Yunhan Xu,
  Phil Chen, Andy Coenen, Clemens Meyer, Katerina Tsihlas, Ada Ma, Juraj
  Gottweis, Jinwei Xing, Chenjie Gu, Jin Miao, Christian Frank, Zeynep Cankara,
  Sanjay Ganapathy, Ishita Dasgupta, Steph Hughes-Fitt, Heng Chen, David Reid,
  Keran Rong, Hongmin Fan, Joost van Amersfoort, Vincent Zhuang, Aaron Cohen,
  Shixiang~Shane Gu, Anhad Mohananey, Anastasija Ilic, Taylor Tobin, John
  Wieting, Anna Bortsova, Phoebe Thacker, Emma Wang, Emily Caveness, Justin
  Chiu, Eren Sezener, Alex Kaskasoli, Steven Baker, Katie Millican, Mohamed
  Elhawaty, Kostas Aisopos, Carl Lebsack, Nathan Byrd, Hanjun Dai, Wenhao Jia,
  Matthew Wiethoff, Elnaz Davoodi, Albert Weston, Lakshman Yagati, Arun Ahuja,
  Isabel Gao, Golan Pundak, Susan Zhang, Michael Azzam, Khe~Chai Sim, Sergi
  Caelles, James Keeling, Abhanshu Sharma, Andy Swing, YaGuang Li, Chenxi Liu,
  Carrie~Grimes Bostock, Yamini Bansal, Zachary Nado, Ankesh Anand, Josh
  Lipschultz, Abhijit Karmarkar, Lev Proleev, Abe Ittycheriah, Soheil~Hassas
  Yeganeh, George Polovets, Aleksandra Faust, Jiao Sun, Alban Rrustemi, Pen Li,
  Rakesh Shivanna, Jeremiah Liu, Chris Welty, Federico Lebron, Anirudh
  Baddepudi, Sebastian Krause, Emilio Parisotto, Radu Soricut, Zheng Xu, Dawn
  Bloxwich, Melvin Johnson, Behnam Neyshabur, Justin Mao-Jones, Renshen Wang,
  Vinay Ramasesh, Zaheer Abbas, Arthur Guez, Constant Segal, Duc~Dung Nguyen,
  James Svensson, Le Hou, Sarah York, Kieran Milan, Sophie Bridgers, Wiktor
  Gworek, Marco Tagliasacchi, James Lee-Thorp, Michael Chang, Alexey Guseynov,
  Ale~Jakse Hartman, Michael Kwong, Ruizhe Zhao, Sheleem Kashem, Elizabeth
  Cole, Antoine Miech, Richard Tanburn, Mary Phuong, Filip Pavetic, Sebastien
  Cevey, Ramona Comanescu, Richard Ives, Sherry Yang, Cosmo Du, Bo Li, Zizhao
  Zhang, Mariko Iinuma, Clara~Huiyi Hu, Aurko Roy, Shaan Bijwadia, Zhenkai Zhu,
  Danilo Martins, Rachel Saputro, Anita Gergely, Steven Zheng, Dawei Jia,
  Ioannis Antonoglou, Adam Sadovsky, Shane Gu, Yingying Bi, Alek Andreev, Sina
  Samangooei, Mina Khan, Tomas Kocisky, Angelos Filos, Chintu Kumar, Colton
  Bishop, Adams Yu, Sarah Hodkinson, Sid Mittal, Premal Shah, Alexandre
  Moufarek, Yong Cheng, Adam Bloniarz, Jaehoon Lee, Pedram Pejman, Paul Michel,
  Stephen Spencer, Vladimir Feinberg, Xuehan Xiong, Nikolay Savinov, Charlotte
  Smith, Siamak Shakeri, Dustin Tran, Mary Chesus, Bernd Bohnet, George Tucker,
  Tamara von Glehn, Carrie Muir, Yiran Mao, Hideto Kazawa, Ambrose Slone, Kedar
  Soparkar, Disha Shrivastava, James Cobon-Kerr, Michael Sharman, Jay
  Pavagadhi, Carlos Araya, Karolis Misiunas, Nimesh Ghelani, Michael Laskin,
  David Barker, Qiujia Li, Anton Briukhov, Neil Houlsby, Mia Glaese, Balaji
  Lakshminarayanan, Nathan Schucher, Yunhao Tang, Eli Collins, Hyeontaek Lim,
  Fangxiaoyu Feng, Adria Recasens, Guangda Lai, Alberto Magni, Nicola~De Cao,
  Aditya Siddhant, Zoe Ashwood, Jordi Orbay, Mostafa Dehghani, Jenny Brennan,
  Yifan He, Kelvin Xu, Yang Gao, Carl Saroufim, James Molloy, Xinyi Wu, Seb
  Arnold, Solomon Chang, Julian Schrittwieser, Elena Buchatskaya, Soroush
  Radpour, Martin Polacek, Skye Giordano, Ankur Bapna, Simon Tokumine, Vincent
  Hellendoorn, Thibault Sottiaux, Sarah Cogan, Aliaksei Severyn, Mohammad
  Saleh, Shantanu Thakoor, Laurent Shefey, Siyuan Qiao, Meenu Gaba, Shuo yiin
  Chang, Craig Swanson, Biao Zhang, Benjamin Lee, Paul~Kishan Rubenstein, Gan
  Song, Tom Kwiatkowski, Anna Koop, Ajay Kannan, David Kao, Parker Schuh, Axel
  Stjerngren, Golnaz Ghiasi, Gena Gibson, Luke Vilnis, Ye Yuan, Felipe~Tiengo
  Ferreira, Aishwarya Kamath, Ted Klimenko, Ken Franko, Kefan Xiao, Indro
  Bhattacharya, Miteyan Patel, Rui Wang, Alex Morris, Robin Strudel, Vivek
  Sharma, Peter Choy, Sayed~Hadi Hashemi, Jessica Landon, Mara Finkelstein,
  Priya Jhakra, Justin Frye, Megan Barnes, Matthew Mauger, Dennis Daun, Khuslen
  Baatarsukh, Matthew Tung, Wael Farhan, Henryk Michalewski, Fabio Viola, Felix
  de Chaumont~Quitry, Charline~Le Lan, Tom Hudson, Qingze Wang, Felix Fischer,
  Ivy Zheng, Elspeth White, Anca Dragan, Jean baptiste Alayrac, Eric Ni,
  Alexander Pritzel, Adam Iwanicki, Michael Isard, Anna Bulanova, Lukas Zilka,
  Ethan Dyer, Devendra Sachan, Srivatsan Srinivasan, Hannah Muckenhirn,
  Honglong Cai, Amol Mandhane, Mukarram Tariq, Jack~W. Rae, Gary Wang, Kareem
  Ayoub, Nicholas FitzGerald, Yao Zhao, Woohyun Han, Chris Alberti, Dan
  Garrette, Kashyap Krishnakumar, Mai Gimenez, Anselm Levskaya, Daniel Sohn,
  Josip Matak, Inaki Iturrate, Michael~B. Chang, Jackie Xiang, Yuan Cao,
  Nishant Ranka, Geoff Brown, Adrian Hutter, Vahab Mirrokni, Nanxin Chen,
  Kaisheng Yao, Zoltan Egyed, Francois Galilee, Tyler Liechty, Praveen
  Kallakuri, Evan Palmer, Sanjay Ghemawat, Jasmine Liu, David Tao, Chloe
  Thornton, Tim Green, Mimi Jasarevic, Sharon Lin, Victor Cotruta, Yi-Xuan Tan,
  Noah Fiedel, Hongkun Yu, Ed Chi, Alexander Neitz, Jens Heitkaemper, Anu
  Sinha, Denny Zhou, Yi Sun, Charbel Kaed, Brice Hulse, Swaroop Mishra, Maria
  Georgaki, Sneha Kudugunta, Clement Farabet, Izhak Shafran, Daniel Vlasic,
  Anton Tsitsulin, Rajagopal Ananthanarayanan, Alen Carin, Guolong Su, Pei Sun,
  Shashank V, Gabriel Carvajal, Josef Broder, Iulia Comsa, Alena Repina,
  William Wong, Warren~Weilun Chen, Peter Hawkins, Egor Filonov, Lucia Loher,
  Christoph Hirnschall, Weiyi Wang, Jingchen Ye, Andrea Burns, Hardie Cate,
  Diana~Gage Wright, Federico Piccinini, Lei Zhang, Chu-Cheng Lin, Ionel Gog,
  Yana Kulizhskaya, Ashwin Sreevatsa, Shuang Song, Luis~C. Cobo, Anand Iyer,
  Chetan Tekur, Guillermo Garrido, Zhuyun Xiao, Rupert Kemp, Huaixiu~Steven
  Zheng, Hui Li, Ananth Agarwal, Christel Ngani, Kati Goshvadi, Rebeca
  Santamaria-Fernandez, Wojciech Fica, Xinyun Chen, Chris Gorgolewski, Sean
  Sun, Roopal Garg, Xinyu Ye, S.~M.~Ali Eslami, Nan Hua, Jon Simon, Pratik
  Joshi, Yelin Kim, Ian Tenney, Sahitya Potluri, Lam~Nguyen Thiet, Quan Yuan,
  Florian Luisier, Alexandra Chronopoulou, Salvatore Scellato, Praveen
  Srinivasan, Minmin Chen, Vinod Koverkathu, Valentin Dalibard, Yaming Xu,
  Brennan Saeta, Keith Anderson, Thibault Sellam, Nick Fernando, Fantine Huot,
  Junehyuk Jung, Mani Varadarajan, Michael Quinn, Amit Raul, Maigo Le, Ruslan
  Habalov, Jon Clark, Komal Jalan, Kalesha Bullard, Achintya Singhal, Thang
  Luong, Boyu Wang, Sujeevan Rajayogam, Julian Eisenschlos, Johnson Jia, Daniel
  Finchelstein, Alex Yakubovich, Daniel Balle, Michael Fink, Sameer Agarwal,
  Jing Li, Dj Dvijotham, Shalini Pal, Kai Kang, Jaclyn Konzelmann, Jennifer
  Beattie, Olivier Dousse, Diane Wu, Remi Crocker, Chen Elkind,
  Siddhartha~Reddy Jonnalagadda, Jong Lee, Dan Holtmann-Rice, Krystal
  Kallarackal, Rosanne Liu, Denis Vnukov, Neera Vats, Luca Invernizzi, Mohsen
  Jafari, Huanjie Zhou, Lilly Taylor, Jennifer Prendki, Marcus Wu, Tom Eccles,
  Tianqi Liu, Kavya Kopparapu, Francoise Beaufays, Christof Angermueller,
  Andreea Marzoca, Shourya Sarcar, Hilal Dib, Jeff Stanway, Frank Perbet, Nejc
  Trdin, Rachel Sterneck, Andrey Khorlin, Dinghua Li, Xihui Wu, Sonam Goenka,
  David Madras, Sasha Goldshtein, Willi Gierke, Tong Zhou, Yaxin Liu, Yannie
  Liang, Anais White, Yunjie Li, Shreya Singh, Sanaz Bahargam, Mark Epstein,
  Sujoy Basu, Li Lao, Adnan Ozturel, Carl Crous, Alex Zhai, Han Lu, Zora Tung,
  Neeraj Gaur, Alanna Walton, Lucas Dixon, Ming Zhang, Amir Globerson, Grant
  Uy, Andrew Bolt, Olivia Wiles, Milad Nasr, Ilia Shumailov, Marco Selvi,
  Francesco Piccinno, Ricardo Aguilar, Sara McCarthy, Misha Khalman, Mrinal
  Shukla, Vlado Galic, John Carpenter, Kevin Villela, Haibin Zhang, Harry
  Richardson, James Martens, Matko Bosnjak, Shreyas~Rammohan Belle, Jeff
  Seibert, Mahmoud Alnahlawi, Brian McWilliams, Sankalp Singh, Annie Louis, Wen
  Ding, Dan Popovici, Lenin Simicich, Laura Knight, Pulkit Mehta, Nishesh
  Gupta, Chongyang Shi, Saaber Fatehi, Jovana Mitrovic, Alex Grills, Joseph
  Pagadora, Dessie Petrova, Danielle Eisenbud, Zhishuai Zhang, Damion Yates,
  Bhavishya Mittal, Nilesh Tripuraneni, Yannis Assael, Thomas Brovelli, Prateek
  Jain, Mihajlo Velimirovic, Canfer Akbulut, Jiaqi Mu, Wolfgang Macherey, Ravin
  Kumar, Jun Xu, Haroon Qureshi, Gheorghe Comanici, Jeremy Wiesner, Zhitao
  Gong, Anton Ruddock, Matthias Bauer, Nick Felt, Anirudh GP, Anurag Arnab,
  Dustin Zelle, Jonas Rothfuss, Bill Rosgen, Ashish Shenoy, Bryan Seybold,
  Xinjian Li, Jayaram Mudigonda, Goker Erdogan, Jiawei Xia, Jiri Simsa, Andrea
  Michi, Yi Yao, Christopher Yew, Steven Kan, Isaac Caswell, Carey Radebaugh,
  Andre Elisseeff, Pedro Valenzuela, Kay McKinney, Kim Paterson, Albert Cui,
  Eri Latorre-Chimoto, Solomon Kim, William Zeng, Ken Durden, Priya Ponnapalli,
  Tiberiu Sosea, Christopher~A. Choquette-Choo, James Manyika, Brona Robenek,
  Harsha Vashisht, Sebastien Pereira, Hoi Lam, Marko Velic, Denese
  Owusu-Afriyie, Katherine Lee, Tolga Bolukbasi, Alicia Parrish, Shawn Lu, Jane
  Park, Balaji Venkatraman, Alice Talbert, Lambert Rosique, Yuchung Cheng,
  Andrei Sozanschi, Adam Paszke, Praveen Kumar, Jessica Austin, Lu Li, Khalid
  Salama, Wooyeol Kim, Nandita Dukkipati, Anthony Baryshnikov, Christos
  Kaplanis, XiangHai Sheng, Yuri Chervonyi, Caglar Unlu, Diego de Las~Casas,
  Harry Askham, Kathryn Tunyasuvunakool, Felix Gimeno, Siim Poder, Chester
  Kwak, Matt Miecnikowski, Vahab Mirrokni, Alek Dimitriev, Aaron Parisi, Dangyi
  Liu, Tomy Tsai, Toby Shevlane, Christina Kouridi, Drew Garmon, Adrian
  Goedeckemeyer, Adam~R. Brown, Anitha Vijayakumar, Ali Elqursh, Sadegh
  Jazayeri, Jin Huang, Sara~Mc Carthy, Jay Hoover, Lucy Kim, Sandeep Kumar, Wei
  Chen, Courtney Biles, Garrett Bingham, Evan Rosen, Lisa Wang, Qijun Tan,
  David Engel, Francesco Pongetti, Dario de Cesare, Dongseong Hwang, Lily Yu,
  Jennifer Pullman, Srini Narayanan, Kyle Levin, Siddharth Gopal, Megan Li,
  Asaf Aharoni, Trieu Trinh, Jessica Lo, Norman Casagrande, Roopali Vij, Loic
  Matthey, Bramandia Ramadhana, Austin Matthews, CJ Carey, Matthew Johnson,
  Kremena Goranova, Rohin Shah, Shereen Ashraf, Kingshuk Dasgupta, Rasmus
  Larsen, Yicheng Wang, Manish~Reddy Vuyyuru, Chong Jiang, Joana Ijazi, Kazuki
  Osawa, Celine Smith, Ramya~Sree Boppana, Taylan Bilal, Yuma Koizumi, Ying Xu,
  Yasemin Altun, Nir Shabat, Ben Bariach, Alex Korchemniy, Kiam Choo, Olaf
  Ronneberger, Chimezie Iwuanyanwu, Shubin Zhao, David Soergel, Cho-Jui Hsieh,
  Irene Cai, Shariq Iqbal, Martin Sundermeyer, Zhe Chen, Elie Bursztein,
  Chaitanya Malaviya, Fadi Biadsy, Prakash Shroff, Inderjit Dhillon, Tejasi
  Latkar, Chris Dyer, Hannah Forbes, Massimo Nicosia, Vitaly Nikolaev, Somer
  Greene, Marin Georgiev, Pidong Wang, Nina Martin, Hanie Sedghi, John Zhang,
  Praseem Banzal, Doug Fritz, Vikram Rao, Xuezhi Wang, Jiageng Zhang, Viorica
  Patraucean, Dayou Du, Igor Mordatch, Ivan Jurin, Lewis Liu, Ayush Dubey, Abhi
  Mohan, Janek Nowakowski, Vlad-Doru Ion, Nan Wei, Reiko Tojo, Maria~Abi Raad,
  Drew~A. Hudson, Vaishakh Keshava, Shubham Agrawal, Kevin Ramirez, Zhichun Wu,
  Hoang Nguyen, Ji Liu, Madhavi Sewak, Bryce Petrini, DongHyun Choi, Ivan
  Philips, Ziyue Wang, Ioana Bica, Ankush Garg, Jarek Wilkiewicz, Priyanka
  Agrawal, Xiaowei Li, Danhao Guo, Emily Xue, Naseer Shaik, Andrew Leach,
  Sadh~MNM Khan, Julia Wiesinger, Sammy Jerome, Abhishek Chakladar,
  Alek~Wenjiao Wang, Tina Ornduff, Folake Abu, Alireza Ghaffarkhah, Marcus
  Wainwright, Mario Cortes, Frederick Liu, Joshua Maynez, Andreas Terzis, Pouya
  Samangouei, Riham Mansour, Tomasz Kępa, François-Xavier Aubet, Anton
  Algymr, Dan Banica, Agoston Weisz, Andras Orban, Alexandre Senges, Ewa
  Andrejczuk, Mark Geller, Niccolo~Dal Santo, Valentin Anklin, Majd~Al Merey,
  Martin Baeuml, Trevor Strohman, Junwen Bai, Slav Petrov, Yonghui Wu, Demis
  Hassabis, Koray Kavukcuoglu, Jeffrey Dean, and Oriol Vinyals.
\newblock {Gemini 1.5: Unlocking Multimodal Understanding Across Millions of
  Tokens of Context}, 2024{\natexlab{a}}.

\bibitem[Team et~al.(2024{\natexlab{b}})Team, Mesnard, Hardin, Dadashi,
  Bhupatiraju, Pathak, Sifre, Rivière, Kale, Love, Tafti, Hussenot, Sessa,
  Chowdhery, Roberts, Barua, Botev, Castro-Ros, Slone, Héliou, Tacchetti,
  Bulanova, Paterson, Tsai, Shahriari, Lan, Choquette-Choo, Crepy, Cer,
  Ippolito, Reid, Buchatskaya, Ni, Noland, Yan, Tucker, Muraru,
  Rozhdestvenskiy, Michalewski, Tenney, Grishchenko, Austin, Keeling,
  Labanowski, Lespiau, Stanway, Brennan, Chen, Ferret, Chiu, Mao-Jones, Lee,
  Yu, Millican, Sjoesund, Lee, Dixon, Reid, Mikuła, Wirth, Sharman, Chinaev,
  Thain, Bachem, Chang, Wahltinez, Bailey, Michel, Yotov, Chaabouni, Comanescu,
  Jana, Anil, McIlroy, Liu, Mullins, Smith, Borgeaud, Girgin, Douglas, Pandya,
  Shakeri, De, Klimenko, Hennigan, Feinberg, Stokowiec, hui Chen, Ahmed, Gong,
  Warkentin, Peran, Giang, Farabet, Vinyals, Dean, Kavukcuoglu, Hassabis,
  Ghahramani, Eck, Barral, Pereira, Collins, Joulin, Fiedel, Senter, Andreev,
  and Kenealy]{gemma}
Gemma Team, Thomas Mesnard, Cassidy Hardin, Robert Dadashi, Surya Bhupatiraju,
  Shreya Pathak, Laurent Sifre, Morgane Rivière, Mihir~Sanjay Kale, Juliette
  Love, Pouya Tafti, Léonard Hussenot, Pier~Giuseppe Sessa, Aakanksha
  Chowdhery, Adam Roberts, Aditya Barua, Alex Botev, Alex Castro-Ros, Ambrose
  Slone, Amélie Héliou, Andrea Tacchetti, Anna Bulanova, Antonia Paterson,
  Beth Tsai, Bobak Shahriari, Charline~Le Lan, Christopher~A. Choquette-Choo,
  Clément Crepy, Daniel Cer, Daphne Ippolito, David Reid, Elena Buchatskaya,
  Eric Ni, Eric Noland, Geng Yan, George Tucker, George-Christian Muraru,
  Grigory Rozhdestvenskiy, Henryk Michalewski, Ian Tenney, Ivan Grishchenko,
  Jacob Austin, James Keeling, Jane Labanowski, Jean-Baptiste Lespiau, Jeff
  Stanway, Jenny Brennan, Jeremy Chen, Johan Ferret, Justin Chiu, Justin
  Mao-Jones, Katherine Lee, Kathy Yu, Katie Millican, Lars~Lowe Sjoesund, Lisa
  Lee, Lucas Dixon, Machel Reid, Maciej Mikuła, Mateo Wirth, Michael Sharman,
  Nikolai Chinaev, Nithum Thain, Olivier Bachem, Oscar Chang, Oscar Wahltinez,
  Paige Bailey, Paul Michel, Petko Yotov, Rahma Chaabouni, Ramona Comanescu,
  Reena Jana, Rohan Anil, Ross McIlroy, Ruibo Liu, Ryan Mullins, Samuel~L
  Smith, Sebastian Borgeaud, Sertan Girgin, Sholto Douglas, Shree Pandya,
  Siamak Shakeri, Soham De, Ted Klimenko, Tom Hennigan, Vlad Feinberg, Wojciech
  Stokowiec, Yu hui Chen, Zafarali Ahmed, Zhitao Gong, Tris Warkentin, Ludovic
  Peran, Minh Giang, Clément Farabet, Oriol Vinyals, Jeff Dean, Koray
  Kavukcuoglu, Demis Hassabis, Zoubin Ghahramani, Douglas Eck, Joelle Barral,
  Fernando Pereira, Eli Collins, Armand Joulin, Noah Fiedel, Evan Senter, Alek
  Andreev, and Kathleen Kenealy.
\newblock Gemma: Open models based on gemini research and technology,
  2024{\natexlab{b}}.

\bibitem[Xinru~Zheng and Lau(2019)]{zheng-sig19}
Ying~Cao Xinru~Zheng, Xiaotian~Qiao and Rynson~W.H. Lau.
\newblock {Content-aware Generative Modeling of Graphic Design Layouts}.
\newblock \emph{ACM Transactions on Graphics (Proc. of SIGGRAPH 2019)}, 38,
  2019.

\bibitem[Yamaguchi(2021)]{yamaguchi2021canvasvae}
Kota Yamaguchi.
\newblock Canvasvae: Learning to generate vector graphic documents.
\newblock In \emph{Proceedings of the IEEE/CVF International Conference on
  Computer Vision}, pages 5481--5489, 2021.

\bibitem[Yang et~al.(2024)Yang, Luo, Qi, Wu, Shan, and
  Chen]{yang2024posterllava}
Tao Yang, Yingmin Luo, Zhongang Qi, Yang Wu, Ying Shan, and Chang~Wen Chen.
\newblock {PosterLLaVa: Constructing a Unified Multi-modal Layout Generator
  with LLM}.
\newblock \emph{arXiv preprint arXiv:2406.02884}, 2024.

\bibitem[Yu et~al.(2022)Yu, Chen, Chen, Meng, Wu, Josel, Niebles, Xiong, and
  Xu]{yu2022layoutdetr}
Ning Yu, Chia-Chih Chen, Zeyuan Chen, Rui Meng, Gang Wu, Paul Josel,
  Juan~Carlos Niebles, Caiming Xiong, and Ran Xu.
\newblock {LayoutDETR: detection transformer is a good multimodal layout
  designer}.
\newblock \emph{arXiv preprint arXiv:2212.09877}, 2022.

\bibitem[Zhong et~al.(2019)Zhong, Tang, and
  Yepes]{zhong2019publaynetlargestdatasetdocument}
Xu Zhong, Jianbin Tang, and Antonio~Jimeno Yepes.
\newblock {PubLayNet: Largest Dataset Ever for Document Layout Analysis}, 2019.

\bibitem[Zhou et~al.(2022)Zhou, Xu, Ma, Ge, Jiang, and Xu]{zhou2022composition}
Min Zhou, Chenchen Xu, Ye Ma, Tiezheng Ge, Yuning Jiang, and Weiwei Xu.
\newblock {Composition-aware Graphic Layout GAN for Visual-textual Presentation
  Designs}.
\newblock \emph{arXiv preprint arXiv:2205.00303}, 2022.

\bibitem[Zhu et~al.(2024)Zhu, Healey, Zhang, Wang, and Sun]{zhu2024automatic}
Wanrong Zhu, Jennifer Healey, Ruiyi Zhang, William~Yang Wang, and Tong Sun.
\newblock {Automatic Layout Planning for Visually-Rich Documents with
  Instruction-Following Models}.
\newblock \emph{arXiv preprint arXiv:2404.15271}, 2024.

\end{thebibliography}
